\documentclass{article}     
 
\usepackage[utf8]{inputenc} 
\usepackage[T1]{fontenc}    
\usepackage{microtype}      
\usepackage[font=normalsize,labelfont=bf]{caption}     
\usepackage[font=normalsize,labelfont=bf]{subcaption}  
 
\usepackage{PRIMEarxiv}     
 
\usepackage{amsmath}        
\usepackage{amssymb}        
\usepackage{amsfonts}       
\usepackage{nicefrac}       
 
\usepackage{algorithm}      
\usepackage{algpseudocode}  
 
\usepackage{graphicx}       
\graphicspath{{media/}}     
\usepackage{adjustbox}      
\usepackage{tikz}           
 
\usepackage{caption}        
\usepackage{subcaption}     
 
\usepackage{float}          
\usepackage[section]{placeins} 
 
\usepackage{booktabs}       
\usepackage{tabularx}       
\usepackage{multirow}       
\usepackage{array}          
\usepackage{colortbl}       
\usepackage{threeparttable} 
\usepackage{diagbox}        
\usepackage{calc}           
\renewcommand{\tabularxcolumn}[1]{m{#1}}  
\usepackage{xcolor}         
\usepackage{tcolorbox}      
\tcbuselibrary{breakable,skins} 
 
\usepackage{fancyhdr}       
\usepackage{lineno}         
\usepackage{pdflscape}      
 
\usepackage{lipsum}         
\usepackage{blindtext}      
 
\usepackage{url}            
\usepackage{hyperref}       
 
\tcbuselibrary{breakable,skins} 
\usepackage{lineno}             
\usepackage{float}              
\makeatletter
\renewcommand{\section}{%
  \@startsection{section}{1}{\z@}%
    {-2.0ex \@plus -0.5ex \@minus -0.2ex}%
    { 1.5ex \@plus  0.3ex \@minus  0.2ex}%
    {\large\bfseries\raggedright}%
}

\makeatother

\tcbset{                    
    breakable,              
    enhanced                
}

\title{
Cross-Paradigm Evaluation of Gaze-Based Semantic Object Identification for Intelligent Vehicles
 
}
 
\author{
  Penghao Deng$^{1}$, Jidong J. Yang$^{1,*}$, Jiachen Bian$^{1}$ \\
  \\
  $^{1}$Smart Mobility \& Infrastructure Laboratory \\
  College of Engineering, University of Georgia, Athens, GA 30602, USA \\
  \texttt{\{Penghao.Deng, Jidong.Yang, Jiachen.Bian\}@uga.edu} \\
  \\
  $^{*}$Corresponding author\\
}

\begin{document}
\maketitle

\begin{abstract}
Understanding where drivers direct their visual attention during driving, as characterized by gaze behavior, is critical for developing next-generation advanced driver-assistance systems and improving road safety. This paper tackles this challenge as a semantic identification task from the road scenes captured by a vehicle's front-view camera. Specifically, the collocation of gaze points with object semantics is investigated using three distinct vision-based approaches: direct object detection (YOLOv13), segmentation-assisted classification (SAM2 paired with EfficientNetV2 versus YOLOv13), and query-based Vision-Language Models, VLMs (Qwen2.5-VL-7b versus Qwen2.5-VL-32b). The results demonstrate that the direct object detection (YOLOv13) and Qwen2.5-VL-32b significantly outperform other approaches, achieving Macro F1-Scores over 0.84. The large VLM (Qwen2.5-VL-32b), in particular, exhibited superior robustness and performance for identifying small, safety-critical objects such as traffic lights, especially in adverse nighttime conditions. Conversely, the segmentation-assisted paradigm suffers from a ``part-versus-whole'' semantic gap that led to large failure in recall. The results reveal a fundamental trade-off between the real-time efficiency of traditional detectors and the richer contextual understanding and robustness offered by large VLMs. These findings provide critical insights and practical guidance for the design of future human-aware intelligent driver monitoring systems.
\end{abstract}


\section{Introduction}\label{introduction}

Despite decades of technological advancements in vehicle engineering and infrastructure, ensuring road safety remains a paramount global challenge. Each year, road traffic incidents result in a staggering number of fatalities and injuries. The U.S. National Highway Traffic Safety Administration projected 39,345 traffic fatalities in the United States for 2024, a figure that, while representing a slight decrease, remains alarmingly high compared to levels from a decade prior \cite{natio2025EarlyEstimateMotor}. On a global scale, road traffic injuries are the leading cause of death for individuals aged 5 to 29, with an estimated 1.19 million lives lost annually \cite{world2023GlobalStatusReport}. A significant contributor to this persistent crisis is driver inattention, a multifaceted issue encompassing any activity that diverts a driver's focus from the primary task of safely operating a vehicle \cite{ahlst2021EyeTrackingDriver, garci2021AssessmentInfluenceTechnologyBased}. Driver distraction, a subset of inattention, is implicated in a substantial portion of police-reported crashes, with estimates suggesting its involvement in 8\% of fatal crashes and 13\% of injury crashes in 2023 alone \cite{natio2025Distracteddriving2023}.

Naturalistic driving studies have provided compelling quantitative evidence linking inattention to elevated crash risk. Research has shown that engaging in visually or manually complex secondary tasks can increase the near-crash or crash risk by a factor of three compared to attentive driving \cite{klaue2015EffectSecondaryTask}. More critically, a single glance away from the forward roadway lasting more than two seconds has been found to increase this risk by at least twofold \cite{simon2014KeepYourEyes}. The proliferation of in-vehicle infotainment systems and personal mobile devices has exacerbated this problem, introducing potent sources of visual, manual, and cognitive distraction into the cockpit \cite{guo2017effectsagecrash}.

In response to this critical safety issue, the automotive industry and regulatory bodies have championed the development and deployment of Driver Monitoring Systems (DMS). These systems utilize in-cabin sensors and artificial intelligence to assess a driver's state in real time, detecting conditions such as drowsiness, distraction, and impairment \cite{dong2011DriverInattentionMonitoring}. While current DMS technologies are effective at identifying overt signs of inattention, such as prolonged eye closure or cell phone use, the next frontier in vehicle safety lies in achieving a more profound understanding of the driver's situational awareness. This requires moving beyond simple distraction detection to proactively assessing the driver's perception of the dynamic environment.

Central to this endeavor is the analysis of the driver's gaze. A driver's point of gaze is the most direct and powerful indicator of their visual attention, offering a window into their cognitive state, intentions, and awareness of surrounding traffic elements \cite{khan2019GazeEyeTracking, ledez2021ImplementingGazeTracking}. By precisely identifying the semantic object at which a driver is looking, it becomes possible to build a new generation of intelligent, human-centric vehicle systems. The applications of this capability can be extensive and transformative.

First, in the domain of Advanced Driver-Assistance Systems (ADAS), it enables the creation of context-aware warning systems. An ADAS that knows a driver has failed to visually register an approaching pedestrian can issue a timely and critical alert, whereas a system that confirms the driver is already monitoring the hazard can suppress a redundant and potentially annoying warning, thereby reducing alarm fatigue \cite{ledez2021ImplementingGazeTracking}. This represents a paradigm shift from vehicle-centric perception, where the system only knows what it sees, to human-centric perception, where the system understands what the driver sees.

Second, for the development of human-centered autonomous vehicles, analyzing human gaze patterns provides invaluable data for designing more intuitive and trustworthy AI driving policies \cite{feng2024Humancentreddesignnext}. During safety-critical handover events in semi-autonomous vehicles, confirming that the driver's attention is appropriately directed toward the relevant hazard is crucial for ensuring a safe transition of control \cite{haghz2024ClassifyingOlderDrivers, kim2025SustainableRealTimeDriver}. Furthermore, gaze behavior serves as a robust, implicit measure of a driver's trust in the automated system, a key factor in technology acceptance and safe operation \cite{walke2019Gazebehaviourelectrodermal}.

Third, this technology can revolutionize in-vehicle Human-Machine Interfaces (HMIs). HMIs can be designed to adaptively present information in the driver's line of sight or to use subtle cues to guide attention toward critical information, enhancing situational awareness without increasing cognitive load \cite{zhu2024DrivingFutureExploring}.

Finally, for driver behavior analysis, such as in post-accident forensics, driver training, or insurance telematics, identifying gazed objects offers a granular level of insight into a driver's risk perception and decision-making processes that is unattainable with current metrics like speed and acceleration data \cite{winla2019Usingtelematicsdata}.

Formally, the research problem addressed in this paper is point-of-gaze object identification: given an image frame from a vehicle's forward-facing camera and a corresponding time-synchronized gaze coordinate of the driver on the image plane, the task is to determine the semantic class of the object at that gaze point, in other words, what object in the driving scene that the driver pay attention to at a particular time (frame). Despite its clear definition, this task is fraught with significant technical challenges that stem from the dynamic nature of real-world driving.

A primary challenge is the immense variability of driving scenes. The performance of any vision-based system must be robust across a wide spectrum of environmental conditions. Adverse weather, such as rain, snow, and fog, can severely degrade image quality, introducing noise and obscuring object features \cite{kumar2023ObjectDetectionAdverse}. Similarly, lighting conditions vary dramatically, from the bright glare of direct sunlight to the low-light and high-contrast environments of nighttime driving, each posing unique difficulties for perception algorithms \cite{liu2025YOLOFDCLImprovedYOLOv8}. The objects of interest themselves, such as pedestrians, vehicles, and traffic signals, exhibit a vast range of appearances, scales, and orientations. They can be partially occluded by other road users or infrastructure, making their complete and accurate identification a non-trivial problem \cite{tammi2025Quantifyingtrackingquality}.

A more fundamental challenge lies in the inherent ambiguity of a single gaze point. Geometrically, a coordinate is ambiguous. It may fall on a small, distinct object like a distant traffic light, on a specific component of a larger object like the wheel of a truck, or in the empty space between two adjacent vehicles \cite{marda2019ResolvingTargetAmbiguity}. This ambiguity is compounded by a semantic dimension: a driver looking at a car's taillight is semantically attending to the ``car'' as a whole entity, not just the taillight. Any effective system must resolve this part-versus-whole relationship to correctly infer the driver's focus of attention. This technical challenge is deeply intertwined with well-documented phenomena in cognitive psychology, such as ``inattentional blindness'' and ``looked-but-failed-to-see'' (LBFTS) errors \cite{wolfe2022Normalblindnesswhen}. LBFTS incidents occur when a driver's gaze is directed at a hazard, yet they fail to consciously perceive it, often due to a lack of expectation or cognitive overload \cite{hersl2003Lookedbutfailedtoseeerrorstraffic, kenne2013InattentionalBlindnessSimulated}. This underscores that the problem is not merely one of geometric localization but of inferring a latent human cognitive state, where the ``ground truth'' is the driver's semantic intent.

Finally, for any of the aforementioned applications to be viable, particularly those involving real-time warnings or control transitions (e.g., between human driving and automated driving), the entire identification process must be executed with minimal latency. This imposes a strict computational budget on the system, favoring architectures that are both accurate and efficient \cite{hoffm2021RealTimeAdaptiveObject}. The necessity of balancing high accuracy under diverse conditions with the demands of real-time performance makes point-of-gaze object identification a uniquely challenging problem at the intersection of computer vision, human factors, and automotive engineering.

To address the task of point-of-gaze object identification, several distinct methodological paradigms can be adapted from the broader fields of computer vision and scene understanding. These can be broadly categorized into object detection-based, segmentation-assisted, and Vision-Language Model (VLM) based approaches.

\subsection{Object Detection-Based Approaches}

The most direct method involves leveraging mature, off-the-shelf object detectors such as the You Only Look Once (YOLO) family or Region-based Convolutional Neural Networks (R-CNNs) \cite{wang2023DriverAttentionDetection, youss2022Trafficsignclassification}. These models have been extensively applied to traffic scene understanding and driver monitoring, demonstrating high efficiency in the case of single-stage detectors like YOLO and high accuracy from two-stage detectors like Faster R-CNN \cite{maity2021FasterRCNNYOLO}. The operational principle is straightforward: if a driver's gaze coordinate falls within a predicted bounding box, the gazed object is identified by that box's class label. However, this approach suffers from several critical limitations in the context of gaze analysis. First is the ``gaze-in-the-gap'' problem: the method fails entirely if the gaze point lands on the background or between detected objects, providing no information even if the driver is looking at a valid but undetected entity \cite{tonin2023ObjectawareGazeTarget}. Second, standard detectors often struggle with small, distant, or partially occluded objects, which are frequent and often safety-critical in driving scenarios \cite{mirza2023SmallObjectDetection}. Finally, a bounding box provides only a coarse localization, meaning a gaze point could fall near the boundary or a corner of a bounding box, where the gazed pixel belongs to the background rather than the object of interest, leading to an incorrect association.

\subsection{Segmentation-Assisted Classification Approaches}

To address the issue with bounding boxes and achieve more precise localization, a two-stage approach can be employed, first using an image segmentation model to isolate the object at the gaze coordinate, followed by a classification model to identify the resulting image crop. Semantic segmentation networks are adept at providing pixel-level segmentation of driving scenes, delineating drivable areas and common object categories with high fidelity \cite{zhao2024Roadsurfacesemantic}. The recent development of foundation models for segmentation, most notably the Segment Anything Model (SAM), offers a powerful new tool \cite{hazza2024SegmentAnythingReview}. SAM can generate high-quality segmentation masks for arbitrary objects based on point or box prompts, demonstrating remarkable zero-shot generalization capabilities \cite{yuan2024Principlesapplicationsadvancements}. However, this paradigm also has significant drawbacks. A primary limitation is its class-agnostic segmentation. It produces a mask but provides no semantic label, necessitating a second classification stage \cite{ravi2024SAM2Segment}. This two-stage pipeline introduces considerable computational overhead, compromising real-time performance. More fundamentally, this approach is susceptible to the ``part-versus-whole'' problem aforementioned. For instance, if a driver's gaze lands on a car's wheel, SAM may accurately segment only the wheel than the whole car.

\subsection{Vision-Language Model (VLM) Based Approaches}

A nascent but highly promising paradigm involves the use of large Vision-Language Models (VLMs). These models, which are pre-trained on vast datasets of paired images and text, have demonstrated powerful capabilities in joint visual and linguistic reasoning \cite{elhen2025VisionLanguageModelsAutonomous}. For the task of gaze object identification, a VLM can be prompted with the full image, the gaze coordinate, and a natural language query such as, ``What is the object at coordinate (x, y)?''. This reframes the problem as a form of Visual Question Answering (VQA) \cite{keska2025EvaluatingMultimodalVisionLanguage}. Modern open-source VLMs, such as the Qwen-VL series, have shown strong performance in spatial understanding, fine-grained object recognition, and answering complex queries about traffic scenarios \cite{bai2023QwenVLVersatileVisionLanguage}. Their ability to perform visual grounding, linking textual concepts to specific image regions, is particularly relevant \cite{wang2025LearningVisualGrounding}. This end-to-end, query-based approach can inherently reason about context and directly address the semantic nature of the task. However, the application of VLMs to this specific, granular problem remains largely unexplored in the literature.

\subsection{Research Gap and Contributions}

Despite the individual advantages and disadvantages within these distinct paradigms, there is a lack of systematic, comparative studies that evaluate their performance for the specific task of identifying drivers' point-of-gaze object. No unified benchmark exists to assess these fundamentally different paradigms on equal footing, particularly under the diverse and challenging environmental conditions of real-world driving. Furthermore, the potential of modern, large VLMs as a direct, query-based solution against established computer vision pipelines for this granular task remains a critical and unexplored question.

This paper aims to fill this critical research gap by presenting a comprehensive, systematic comparison of different vision-based paradigms for identifying the semantic object at a driver's gaze point. In this study, five distinct methods, derived from the three paradigms previously introduced, were evaluated on a newly developed benchmark designed to test performance in realistic and challenging driving scenarios. The main contributions of this work are as follows:

\begin{enumerate}
\def\labelenumi{\arabic{enumi}.}
\item
  The design and implementation of a comprehensive framework for systematically evaluating and comparing three distinct paradigms (object-detection-based, segmentation-assisted, and VLM-based) for driver gazed object identification.
\item
  The first in-depth investigation into the application of large VLMs (specifically Qwen-VL) for this task, providing a direct performance comparison against state-of-the-art computer vision techniques.
\item
  The development and introduction of a new, manually annotated benchmark dataset based on BDD100K, featuring varied and challenging driving conditions (day/night, clear/rainy) to facilitate robust and realistic evaluation.
\item
  A detailed analysis of the performance trade-offs, strengths, and weaknesses of each method across different environmental scenarios and object categories, offering practical insights for developing future driver monitoring and scene perception systems.
\end{enumerate}

The remainder of this paper is organized as follows. Section 2 details our methodology, covering the benchmark dataset, the compared vision paradigms, and the evaluation metrics. Section 3 presents a systematic analysis of the experimental results. Section 4 discusses the key findings, including the practical implications and the practical trade-off between performance and efficiency. Finally, Section 5 concludes the paper with a summary of our findings and outlines directions for future research.

\section{Benchmark Dataset}\label{benchmark-dataset}

To conduct a rigorous and meaningful evaluation of the different vision-based paradigms, a diverse and realistic benchmark dataset is essential. For this study, a benchmark dataset was created from the BDD100K dataset \cite{yu2020BDD100KDiverseDriving} that has a comprehensive coverage of driving scenarios, which includes a wide variety of weather conditions, times of day, and scene types. These characteristics provide a realistic and challenging basis for assessing the performance and robustness of different methods.

The benchmark dataset was created by selecting images exclusively from the city street scene category. These images were then filtered to construct four distinct and challenging environmental conditions: Clear Daytime, Clear Night, Rainy Daytime, and Rainy Night. The annotation protocol involved a manual process where points were placed on objects of interest within each selected image. This procedure was designed to simulate plausible driver gaze points. The final dataset is composed of a collection of image-coordinate-label triplets, which serve as the ground truth for all subsequent experiments.

\begin{figure}[htbp]
  \centering
  \begin{subfigure}[b]{0.49\linewidth}
    \centering
    \includegraphics[width=\linewidth]{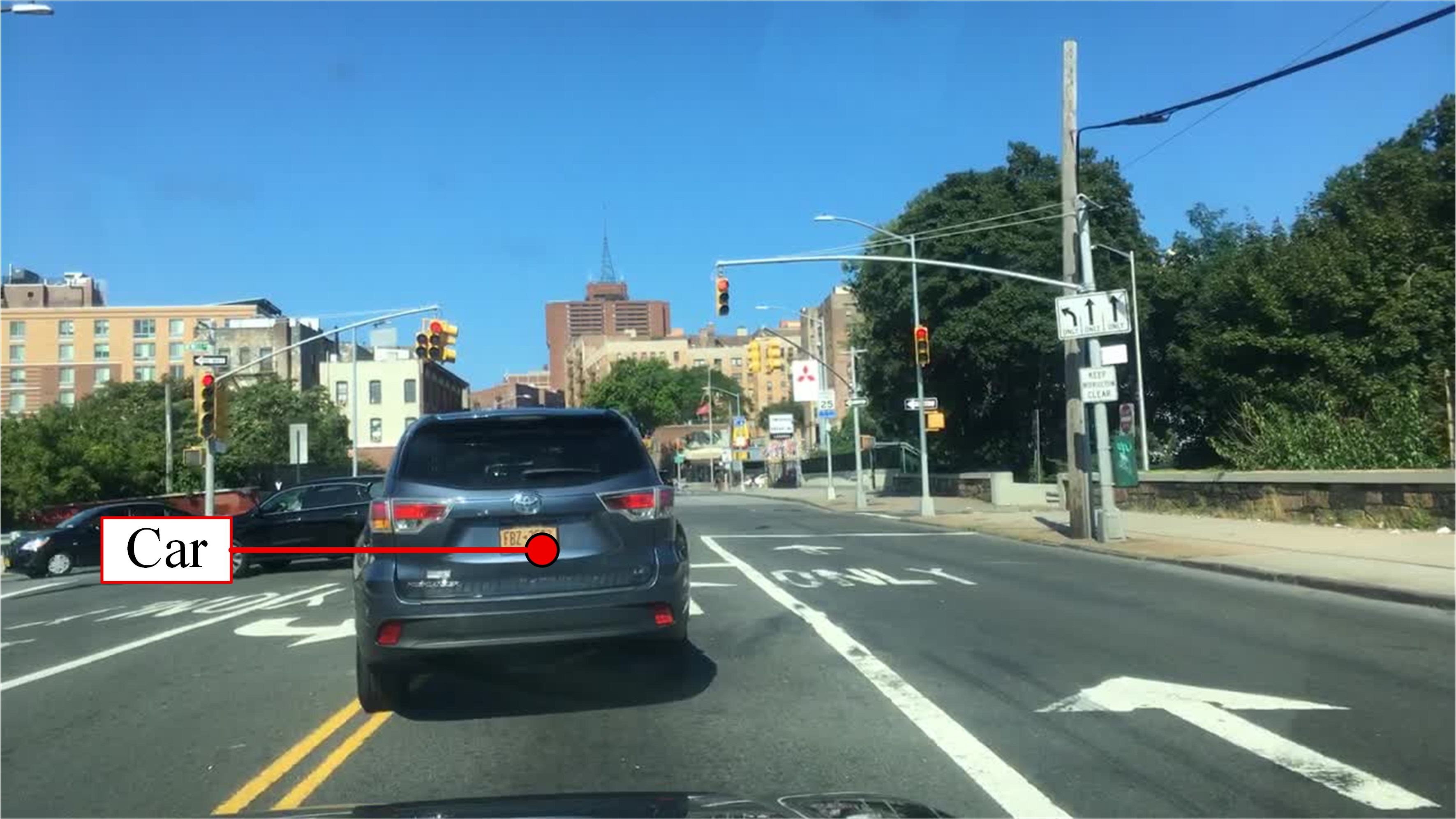}
    \caption{Clear Daytime}
    \label{Figure-1a}
  \end{subfigure}
  \hfill
  \begin{subfigure}[b]{0.49\linewidth}
    \centering
    \includegraphics[width=\linewidth]{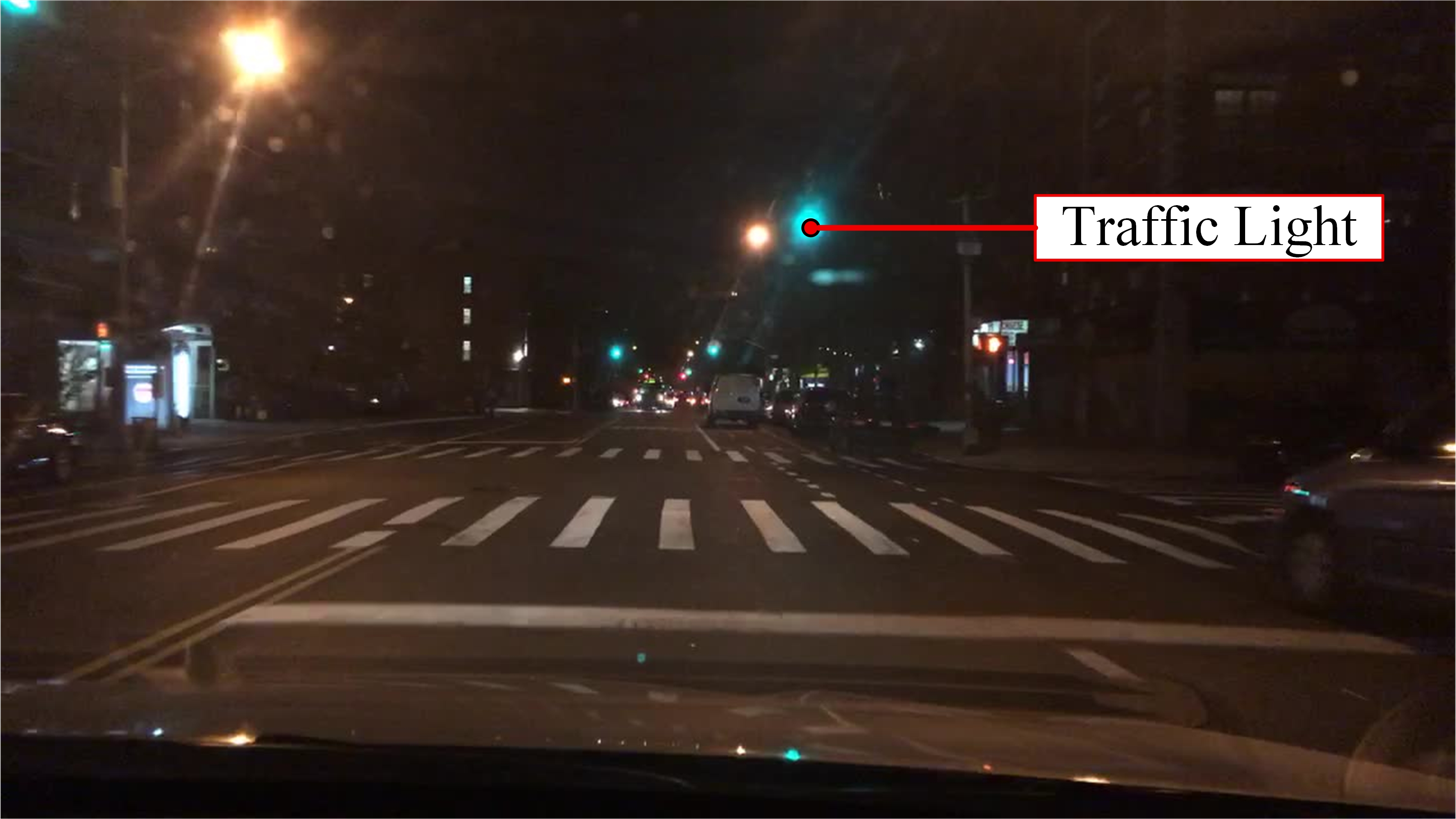}
    \caption{Clear Night}
    \label{Figure-1b}
  \end{subfigure}

  \vspace{1em} 

  \begin{subfigure}[b]{0.49\linewidth}
    \centering
    \includegraphics[width=\linewidth]{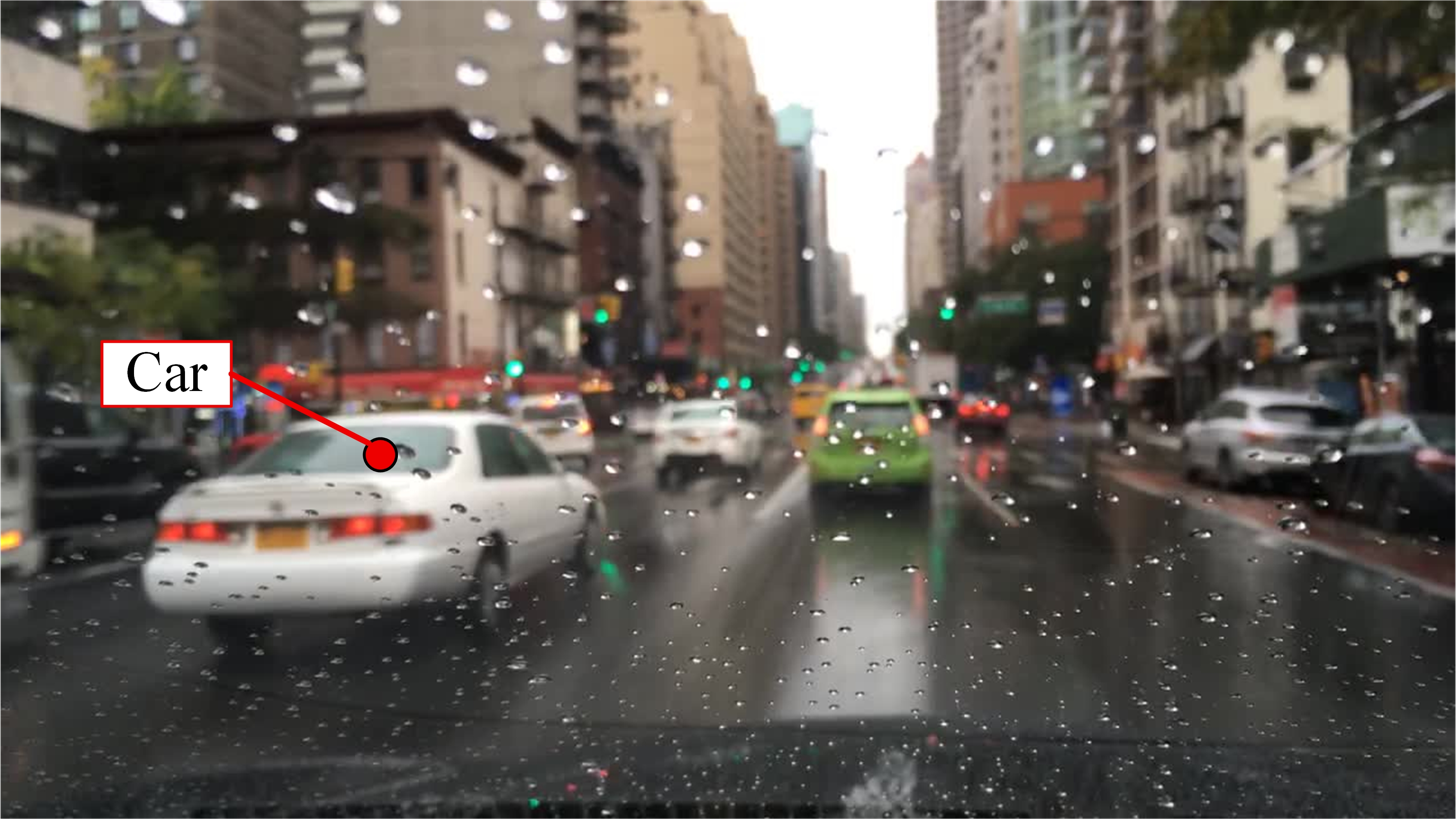}
    \caption{Rainy Daytime}
    \label{Figure-1c}
  \end{subfigure}
  \hfill
  \begin{subfigure}[b]{0.49\linewidth}
    \centering
    \includegraphics[width=\linewidth]{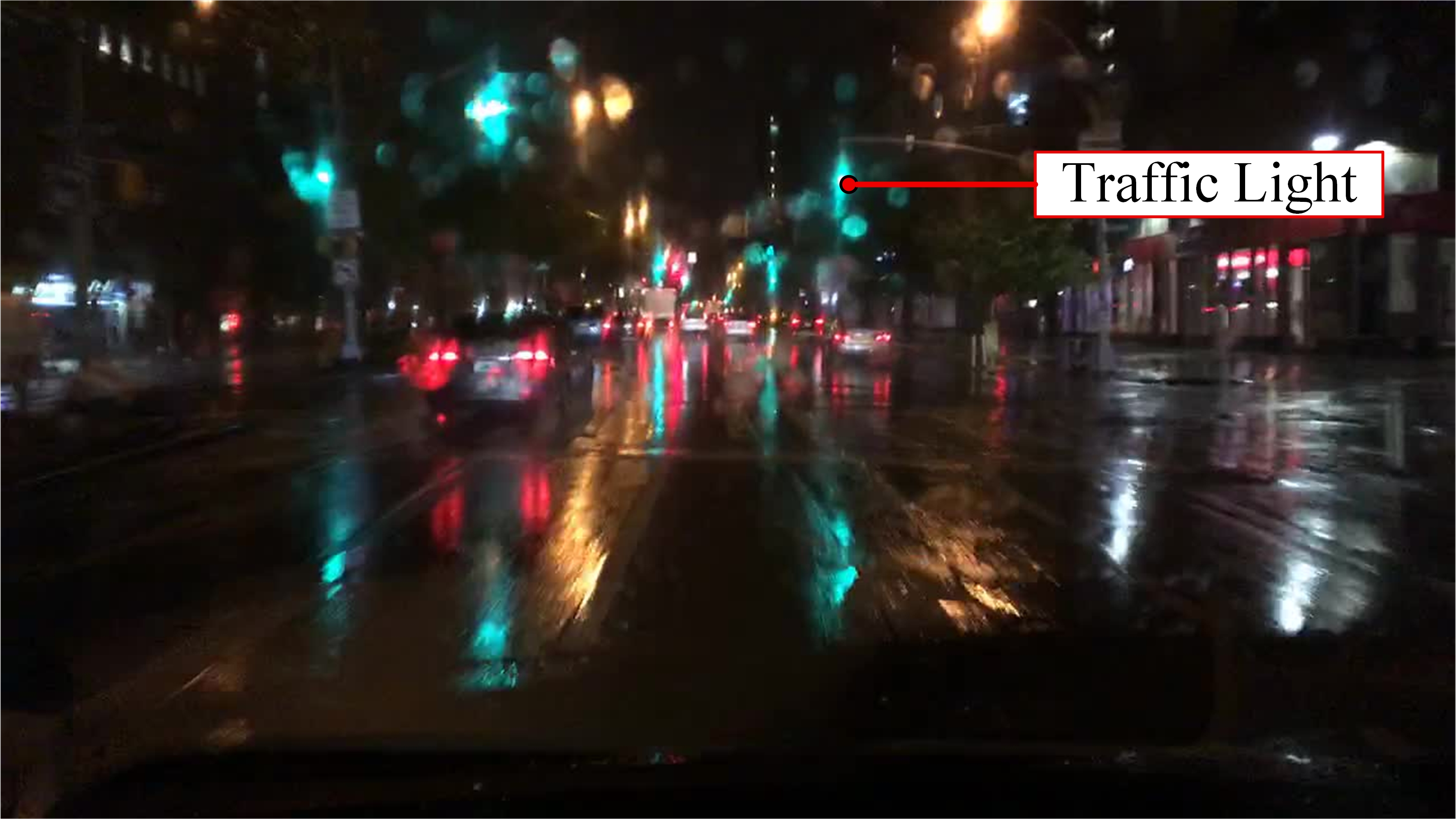}
    \caption{Rainy Night}
    \label{Figure-1d}
  \end{subfigure}

  \caption{Representative examples from the four annotated scenarios and annotation protocol.}
  \label{Figure-1}
\end{figure}

Figure~\ref{Figure-1} presents representative examples from each of the four annotated scenarios, illustrating the annotation method and the visual complexity of the task. Each scenario presents unique challenges to vision algorithms. The Clear Daytime condition, shown in Figure~\ref{Figure-1a}, features good visibility but can include harsh shadows and glare. The Clear Night condition in Figure~\ref{Figure-1b} is characterized by low ambient light and high contrast from artificial sources such as streetlights and vehicle headlights. In the Rainy Daytime scenario, shown in Figure~\ref{Figure-1c}, image quality is degraded by reduced visibility, reflections on wet surfaces, and occlusions from raindrops on the windshield. The Rainy Night scenario, depicted in Figure~\ref{Figure-1d}, is most challenging as it combines the difficulties of nighttime driving with the image degradation effects of rain. The red markers shown in each image are examples of the simulated gaze point annotations used for model evaluation.

Five target object classes were selected for this study: Person, Car, Bus, Truck, and Traffic Light. These classes, corresponding to class IDs 0, 2, 5, 7, and 9, respectively in the original dataset, were chosen based on their prevalence and safety importance in typical urban driving environments. The class IDs align with the COCO dataset \cite{lin2015MicrosoftCOCOCommon}, providing a standardized basis for evaluation. To ensure a fair and consistent assessment across all methods, a label mapping process was required. This was necessary to harmonize the different class definitions used by the pre-trained models, specifically the COCO-based taxonomy of the YOLOv13 object detector \cite{lei2025YOLOv13RealTimeObject} and the ImageNet-1K-based taxonomy of the EfficientNetV2 classifier \cite{tan2021EfficientNetV2SmallerModels}.

Table~\ref{Table-1} details the mapping used to align the ImageNet-1K classes with the five broader target categories. The table shows how multiple, more granular ImageNet-1K classes were consolidated into a single target class. For example, ImageNet-1K classes such as beach wagon, cab, and convertible were all mapped to the Car category. This mapping process introduces a potential challenge for the classifier-based method. The definitions for the Person class in ImageNet-1K, for instance, include categories like ballplayer and groom, which are not representative of typical pedestrians in a driving context and could potentially limit the classifier's performance on that specific class.

\begin{table}[htbp]
  \centering
  \caption{Class Label Mapping.}
  \label{Table-1}
  \begin{tabularx}{\linewidth}{l *{3}{>{\centering\arraybackslash}X}}
    \toprule
    COCO Class ID & COCO Class Name & ImageNet-1K Class ID & ImageNet-1K Class Name \\
    \midrule
    0 & person & 981 & ballplayer \\
    & & 982 & groom \\
    & & 983 & scuba diver \\
    \midrule
    2 & car & 436 & beach wagon \\
    & & 468 & cab \\
    & & 511 & convertible \\
    & & 627 & limousine \\
    & & 661 & Model T \\
    & & 705 & passenger car \\
    & & 751 & racer \\
    & & 817 & sports car \\
    \midrule
    5 & bus & 654 & minibus \\
    & & 779 & school bus \\
    & & 874 & trolleybus \\
    \midrule
    7 & truck & 555 & fire engine \\
    & & 569 & garbage truck \\
    & & 675 & moving van \\
    & & 717 & pickup \\
    & & 864 & tow truck \\
    & & 867 & trailer truck \\
    \midrule
    9 & traffic light & 920 & traffic light \\
    \bottomrule
  \end{tabularx}
\end{table}

The final benchmark dataset consists of 1,355 manually annotated instances distributed across a total of 400 images. To facilitate a balanced comparison of model performance across different conditions, an equal number of images, 100, was selected for each of the four environmental scenarios described previously.

\begin{figure}[htbp]
  \centering
  \begin{adjustbox}{max width=\linewidth}
    \includegraphics{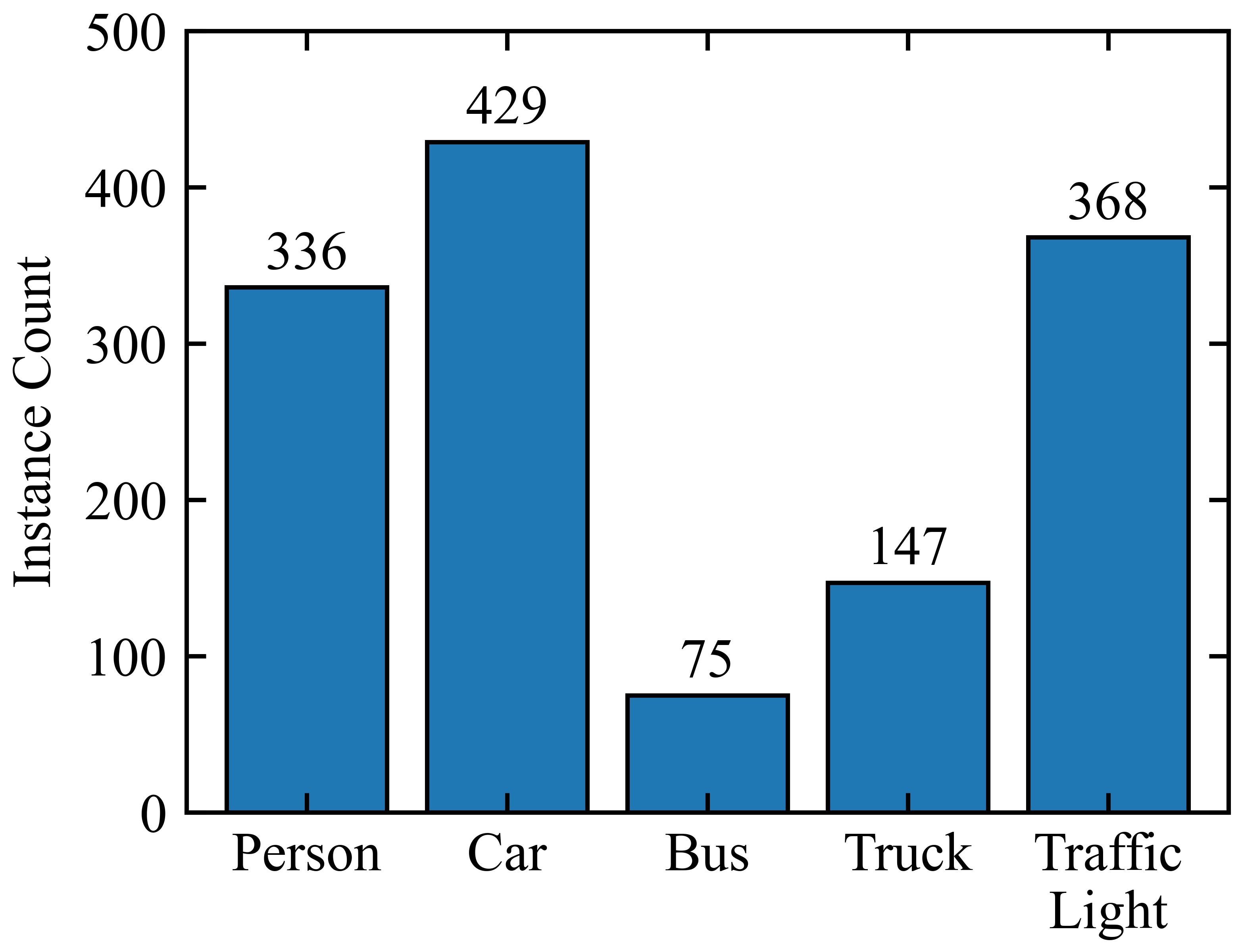}
  \end{adjustbox}
  \caption{Distribution of annotated instances per class.}
  \label{Figure-2}
\end{figure}

Figure~\ref{Figure-2} illustrates the distribution of these annotated instances across the five target classes. The distribution reveals a significant class imbalance, with Car being the most frequent class (429 instances) and Bus being the least frequent (75 instances). This imbalance is considered representative of real-world urban driving, where certain vehicle types are encountered more often than others. This characteristic underscores the importance of using evaluation metrics, such as the Macro F1-Score, that are robust to imbalanced class distributions.

\begin{figure}[htbp]
  \centering
  \begin{subfigure}[b]{0.49\linewidth}
    \centering
    \includegraphics[width=\linewidth]{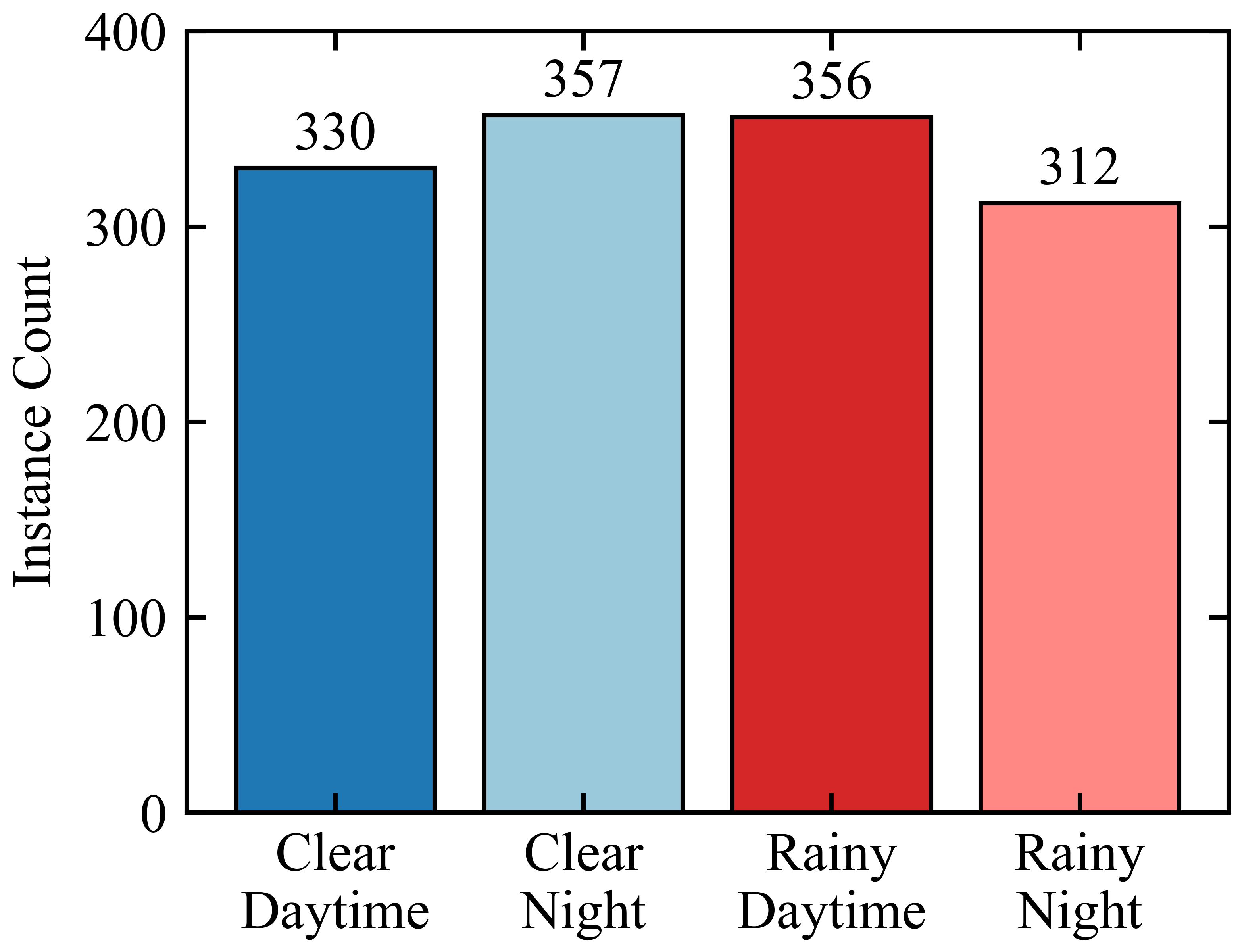}
    \caption{total instance count}
    \label{Figure-3a}
  \end{subfigure}
  \hfill
  \begin{subfigure}[b]{0.49\linewidth}
    \centering
    \includegraphics[width=\linewidth]{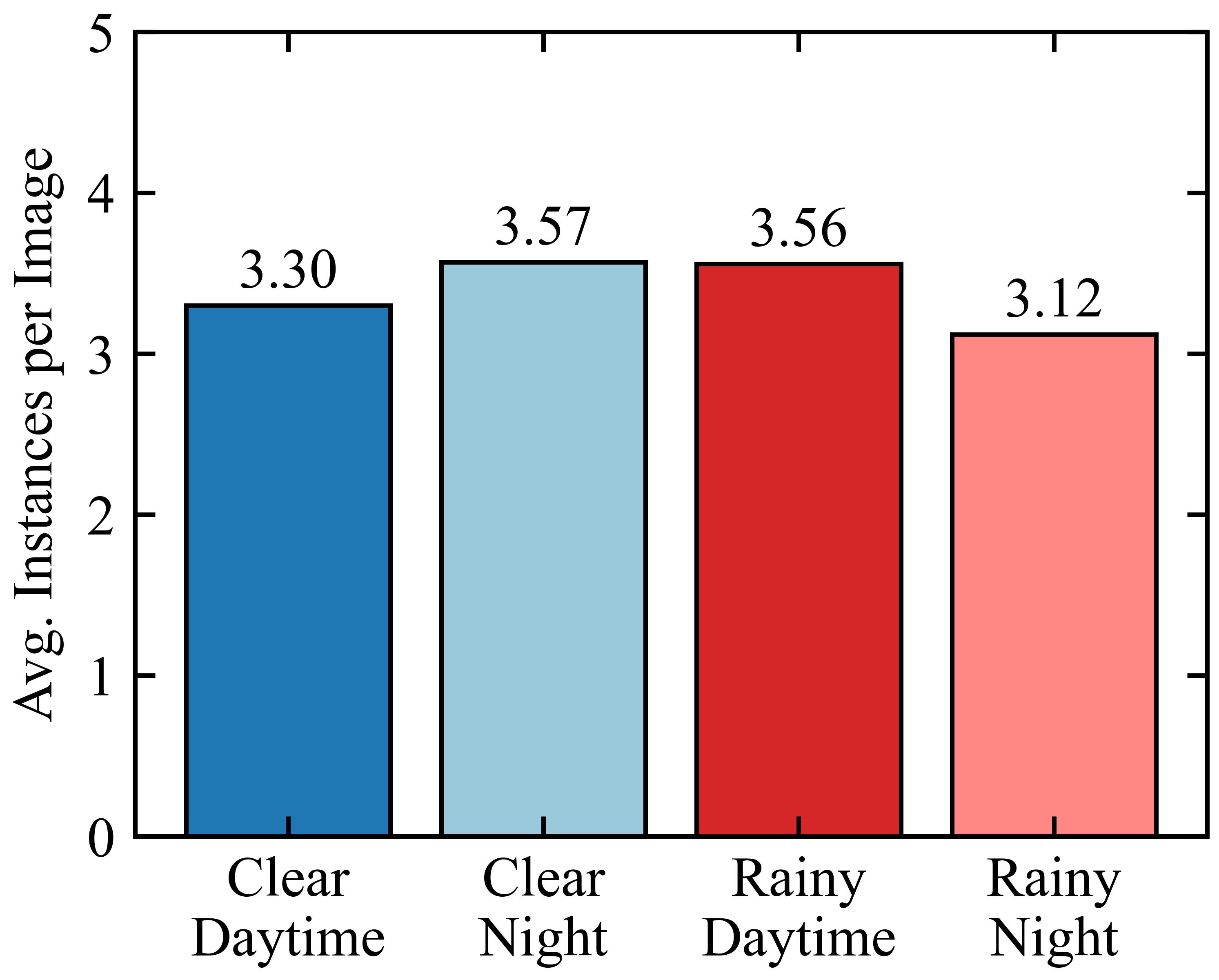}
    \caption{average instance density}
    \label{Figure-3b}
  \end{subfigure}
  \caption{Distribution and density of instances across scenarios.}
  \label{Figure-3}
\end{figure}

A more detailed analysis of the instance distribution and density across the four scenarios is provided in Figure~\ref{Figure-3}. While the number of images per scenario is balanced, the number of annotated objects within them varies. Figure~\ref{Figure-3a} shows the total instance count for each scenario. Figure~\ref{Figure-3b} presents a more nuanced view by showing the average number of instances per image. This analysis reveals that the Clear Night and Rainy Daytime scenarios have the highest object density, with an average of 3.57 and 3.56 instances per image, respectively. This suggests that these scenes are more visually cluttered, which could pose a greater perceptual challenge to the evaluated models.

\section{Methodology}\label{methodology}

\subsection{Comparison of Vision-based Approaches}\label{comparison-of-vision-based-approaches}

This section describes the three paradigms considered in this study and systematically compares five methods: one under Paradigm 1, two under Paradigm 2, and two under Paradigm 3.

\begin{enumerate}
\def\labelenumi{\arabic{enumi}.}
\item
  Paradigm 1 (Object Detection-Based): A direct identification approach utilizing the YOLOv13 model.
\item
  Paradigm 2 (Segmentation-Assisted Classification): A two-stage pipeline evaluated through two specific methods: SAM2 combined with EfficientNetV2 and SAM2 combined with YOLOv13.
\item
  Paradigm 3 (Vision-Language Model-Based): A query-based approach assessed using two model scales (referred to as two methods): the 7-billion parameter version (Qwen2.5-VL-7b) and the 32-billion parameter version (Qwen2.5-VL-32b).
\end{enumerate}

\begin{figure}[htbp]
  \centering
  \begin{subfigure}[b]{\linewidth}
    \centering
    \begin{adjustbox}{max width=\linewidth}
      \includegraphics{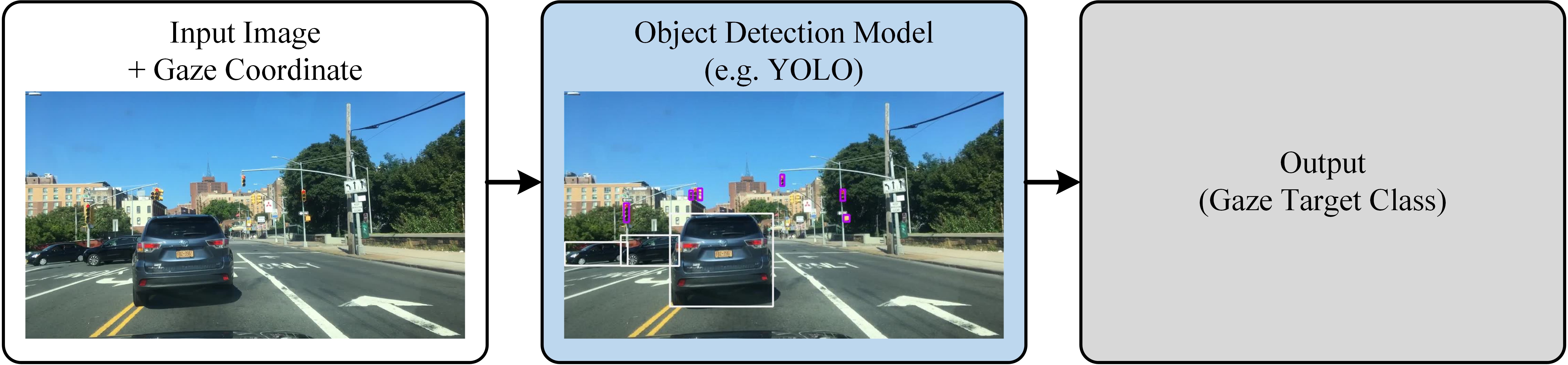}
    \end{adjustbox}
    \caption{Object Detection-Based}
    \label{Figure-4a}
  \end{subfigure}

  \begin{subfigure}[b]{\linewidth}
    \centering
    \begin{adjustbox}{max width=\linewidth}
      \includegraphics{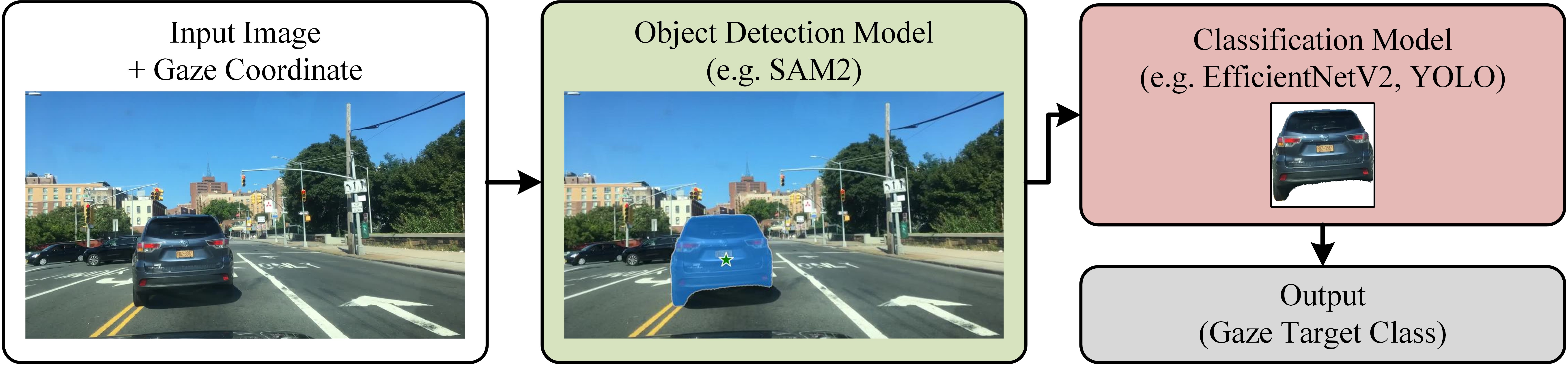}
    \end{adjustbox}
    \caption{Segmentation-Assisted Classification}
    \label{Figure-4b}
  \end{subfigure}

  \begin{subfigure}[b]{\linewidth}
    \centering
    \begin{adjustbox}{max width=\linewidth}
      \includegraphics{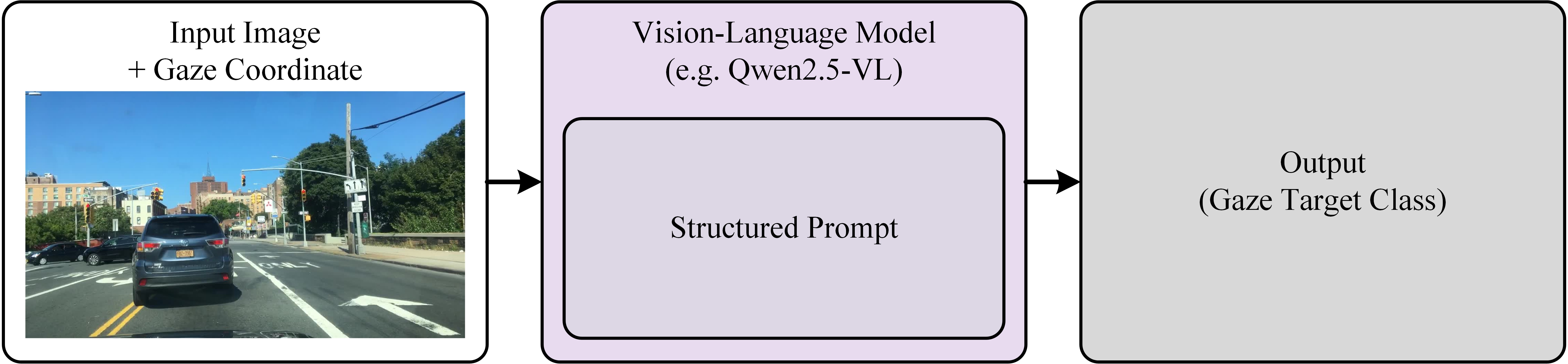}
    \end{adjustbox}
    \caption{Vision-Language Model-Based}
    \label{Figure-4c}
  \end{subfigure}

  \caption{Visualization of the three methodological paradigms.}
  \label{Figure-4}
\end{figure}

Figure~\ref{Figure-4} provides a high-level schematic that visually summarizes the core workflow of each of the three paradigms. The Object Detection-Based paradigm, shown in Figure~\ref{Figure-4a}, is a single-step process where an object detection model processes the input image and coordinate to directly output the class of the object containing the coordinate. The Segmentation-Assisted paradigm, depicted in Figure~\ref{Figure-4b}, is a two-stage pipeline. The first stage uses a segmentation model to isolate the object at the gaze coordinate, and the second stage uses a classification model to identify the resulting crop. The Vision-Language Model paradigm, illustrated in Figure~\ref{Figure-4c}, is an end-to-end process where the input image, coordinate, and a structured text prompt are fed into a VLM, which directly outputs the semantic class.

\subsubsection{Paradigm 1: Object Detection-Based}\label{paradigm-1-object-detection-based}

This paradigm represents the most direct and computationally efficient solution. The specific implementation, Method 1, utilizes a pre-trained YOLOv13 model. The process begins by running the YOLOv13 model on the full input image, which generates a list of predicted bounding boxes, each associated with a class label and a confidence score. The decision logic then checks if the provided gaze coordinate falls within the geometric boundaries of a detected box. If it does, the class label of that box is assigned as the final prediction. A prediction of class ID -1 is recorded under three specific conditions: (1) the gaze coordinate does not fall within any detected bounding box, (2) the confidence score of the bounding box containing the coordinate is below a threshold of 0.4, or (3) the predicted class is not one of the five pre-defined target categories. This value signifies an inability to identify a valid object at the designated point.

\subsubsection{Paradigm 2: Segmentation-Assisted Classification}\label{paradigm-2-segmentation-assisted-classification}

This paradigm aims to achieve higher precision by using pixel-level segmentation to isolate the object of interest prior to classification. The approach involves a two-stage pipeline common to two different methods. In the first stage, the SAM2 model is employed. The driver's gaze coordinate is provided to SAM2 as a positive point prompt, which guides the model to generate a precise pixel-level segmentation mask for the object at that location. Following segmentation, a standardized image preparation step is performed. The object is cropped from the original image based on the generated mask. To create a uniform input for the subsequent classification stage, a square image is constructed. This new image places the cropped object at its center, and all background areas within the square are filled with a neutral white color.

The two methods within this paradigm differ only in the second stage:

\begin{enumerate}
\def\labelenumi{\arabic{enumi}.}
\item
  Method 2-1 (SAM2+EfficientNetV2): The prepared square image of the isolated object is passed to a pre-trained EfficientNetV2 model, which performs the final classification. If the class predicted by EfficientNetV2 does not correspond to one of the five target categories according to the mapping in Table~\ref{Table-1}, the final prediction of class ID is set to -1.
\item
  Method 2-2 (SAM2+YOLOv13): This method serves as a control experiment. The process is identical to Method 2-1, but the classification model is replaced with the YOLOv13 model. A prediction of class ID -1 is assigned if the predicted class the YOLOv13 model is not one of the five target categories. This design allows for an analysis of whether performance limitations in this paradigm originate from the initial segmentation stage or the subsequent classification model.
\end{enumerate}

\subsubsection{Paradigm 3: Vision-Language Model-Based}\label{paradigm-3-vision-language-model-based}

The third paradigm explores the capabilities of large-scale VLMs to solve the task through contextual reasoning. This approach reframes the problem as a Visual Question Answering (VQA) task. The full input image, the (x, y) gaze coordinate, and a carefully structured text prompt are provided as a single, combined input to the VLM. The text prompt, detailed in Figure~\ref{Figure-5}, is designed to explicitly ask the model to identify the object at the given coordinate. Crucially, the prompt also constrains the possible answers to the predefined list of five target classes plus an ``Unsure'' option. This ``Unsure'' category effectively functions as the -1 class, providing a designated output for cases of uncertainty and ensuring the model's responses are directly comparable to the other methods.

\begin{figure}[htbp]
  \centering
  \begin{tcolorbox}[
    sharp corners,   
    colback=white,   
    colframe=black,  
    boxrule=0.5pt,   
    width=\linewidth, 
    boxsep=5pt       
  ]
    \setlength{\parindent}{0pt} 
    
    You are a precise image analysis assistant. \\
    Your task is to identify the object located at a specific pixel coordinate within the provided image. \par
    \vspace{1em}
    
    **Follow these strict rules:** \par
    \vspace{1em}
    
    1. You must select the best-matching category from the following list: '[Person, Car, Bus, Truck, Traffic Light, Unsure]'. \\
    2. Your response must be **strictly** one of the labels from the list (e.g., ``Car''). **Do not** include any additional explanations, descriptions, or extraneous characters. \\
    3. Prioritize the five specific object categories. Use ``Unsure'' only as the last resort when you are completely uncertain. \par
    \vspace{1em}
    
    Now, analyze this image and tell me what object is located at the coordinate '[\{x\}, \{y\}]'.
  \end{tcolorbox}
  \caption{The structured prompt used for the VLM based methods.}
  \label{Figure-5}
\end{figure}

Two methods were evaluated to investigate the impact of model scale on performance:

\begin{enumerate}
\def\labelenumi{\arabic{enumi}.}
\item
  Method 3-1 (Qwen2.5-VL-7b) utilizes the 7-billion parameter version of the Qwen-VL model.
\item
  Method 3-2 (Qwen2.5-VL-32b) utilizes the much larger 32-billion parameter version of the model.
\end{enumerate}

The comparison of these two models is intended to provide insight into how model scale affects performance on this specific and fine-grained visual identification task.

\subsection{Evaluation Metrics}\label{evaluation-metrics}

To quantitatively assess the performance of the compared methodologies, a comprehensive set of standard classification metrics was adopted. The primary metrics used for evaluation were Accuracy, Precision, Recall, and the F1-Score. Accuracy is defined as the ratio of all correct predictions to the total number of instances. For a more detailed analysis of error types, Precision, Recall, and F1-Score were calculated as defined in Equations (1) to (3), respectively.

\begin{gather}
    \text{Precision} = \frac{TP}{TP + FP} \label{eq:precision} \\
    \text{Recall} = \frac{TP}{TP + FN} \label{eq:recall} \\
    \text{F1-Score} = 2 \times \frac{\text{Precision} \cdot \text{Recall}}{\text{Precision} + \text{Recall}} \label{eq:f1}
\end{gather}

where \(TP\), \(FP\), and \(FN\) represent the counts of True Positives, False Positives, and False Negatives for a given class, respectively. Precision measures the model's ability to avoid making false positive predictions, while Recall measures its ability to find all ground truth instances. The F1-Score provides a single, balanced measure of performance by taking the harmonic mean of Precision and Recall.

A special consideration was made for handling model outputs that did not correspond to a target class. A prediction of class ID of ``-1'' signifies a failure by a model to assign one of the five target categories. Such failures can occur due to various method-specific conditions, including low confidence scores, the gaze point falling outside a valid detection region, or an out-of-scope classification. In the evaluation, any instance resulting in a ``-1'' prediction was treated as a false negative for its true ground truth class. This approach correctly penalizes a model's recall for its failure to identify a valid object, as the ground truth dataset contains labels for only the five target object classes.

Furthermore, to account for the significant class imbalance in the dataset, as established in Figure~\ref{Figure-2}, Macro-Averaged metrics were used as the primary indicators of overall performance. Macro-Precision, Macro-Recall, and the Macro F1-Score are calculated by computing the metric independently for each of the five target classes and then taking the unweighted average. This method gives equal weight to each class, regardless of its frequency, thereby providing a more fair and robust assessment of a model's ability to perform well across all object categories.

\section{Results}\label{results}

This section presents a systematic evaluation of the five compared methodologies on the developed benchmark dataset. The analysis begins with an overview of the overall performance, followed by detailed investigations into model robustness across different scenarios and effectiveness on various object categories.

\subsection{Overall Performance Comparison}\label{overall-performance-comparison}

First, the overall performance of the five vision-based methods was evaluated on the benchmark dataset. To ensure a comprehensive and fair assessment in the presence of inherent class imbalance, four evaluation metrics were considered: overall accuracy, macro-precision, macro-recall, and the macro F1-score.

\begin{figure}[htbp]
  \centering
  \begin{subfigure}[b]{0.49\linewidth}
    \centering
    \includegraphics[width=\linewidth]{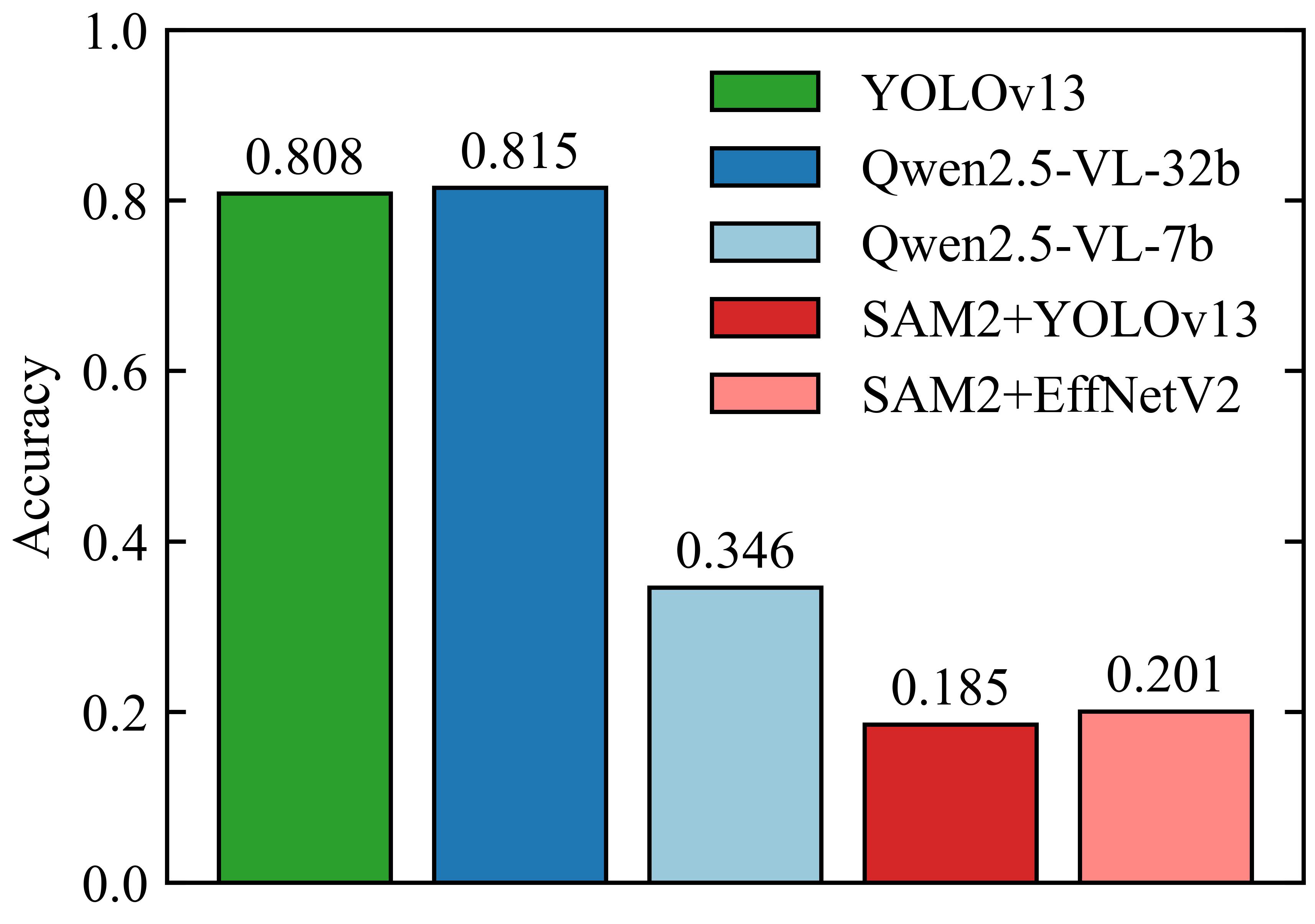}
    \caption{Accuracy}
    \label{Figure-6a}
  \end{subfigure}
  \hfill
  \begin{subfigure}[b]{0.49\linewidth}
    \centering
    \includegraphics[width=\linewidth]{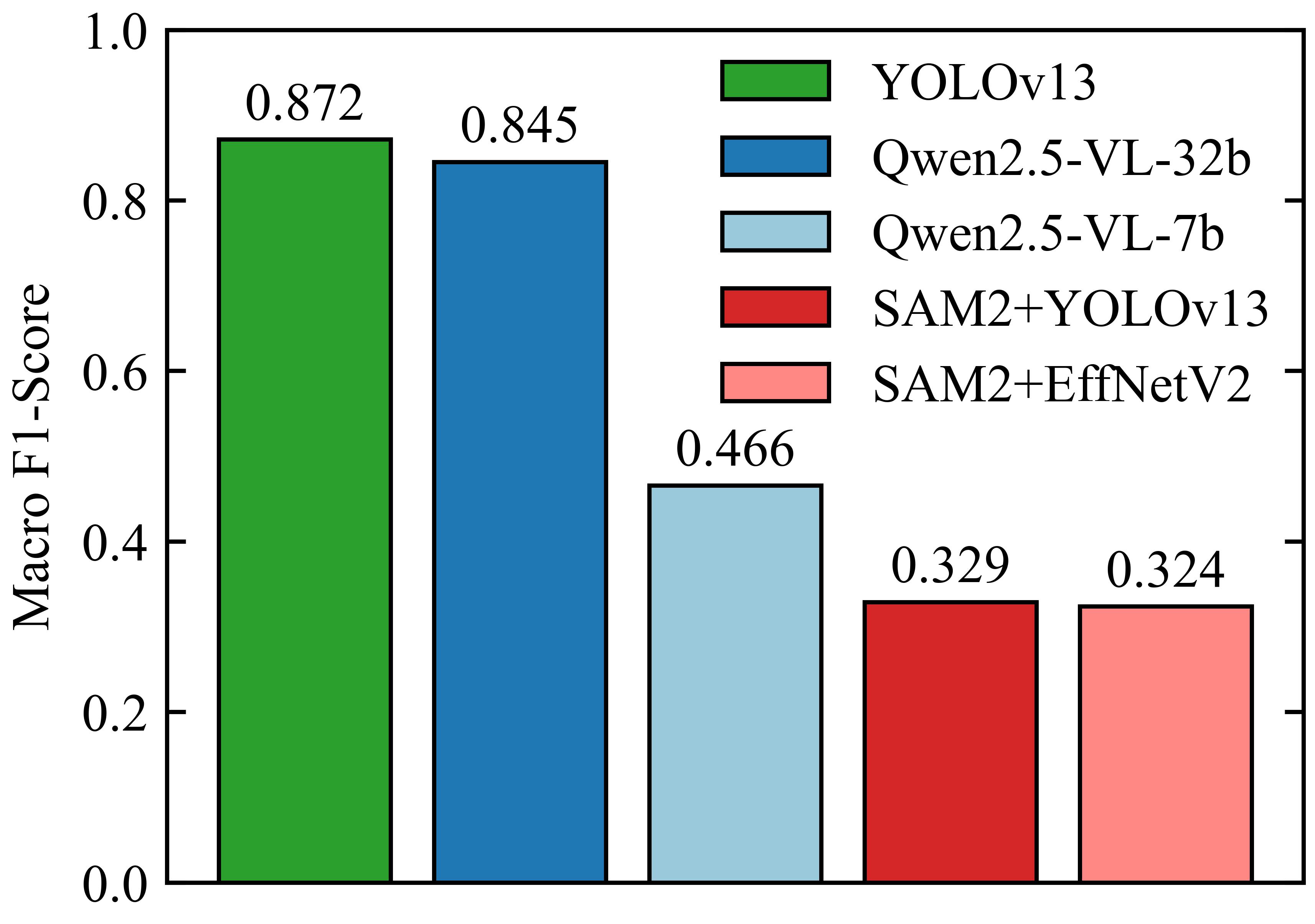}
    \caption{Macro F1-Score}
    \label{Figure-6b}
  \end{subfigure}
  \caption{Overall performance comparison.}
  \label{Figure-6}
\end{figure}

Figure~\ref{Figure-6} presents the primary, high-level findings of the study, comparing the evaluated methods in terms of overall accuracy and the more robust macro F1-score. As shown in Figure~\ref{Figure-6a}, the overall accuracy results reveal clearly the top-two performers. The Qwen2.5-VL-32b model achieved the highest accuracy at 0.815, closely followed by the YOLOv13 model at 0.808. In contrast, the other three methods performed significantly worse, with the two SAM2-based approaches scoring below 0.201. Nonetheless, the overall accuracy can be misleading in the presence of the class imbalance. A more suitable and robust metric for a class-imbalanced dataset is the Macro F1-Score, as plotted in Figure~\ref{Figure-6b}. Using this metric, the YOLOv13 model (0.872) demonstrates a slightly better performance than the Qwen2.5-VL-32b model (0.845), while there is still a large performance gap for the other three methods.

\begin{figure}[htbp]
  \centering
  \begin{subfigure}[b]{0.49\linewidth}
    \centering
    \includegraphics[width=\linewidth]{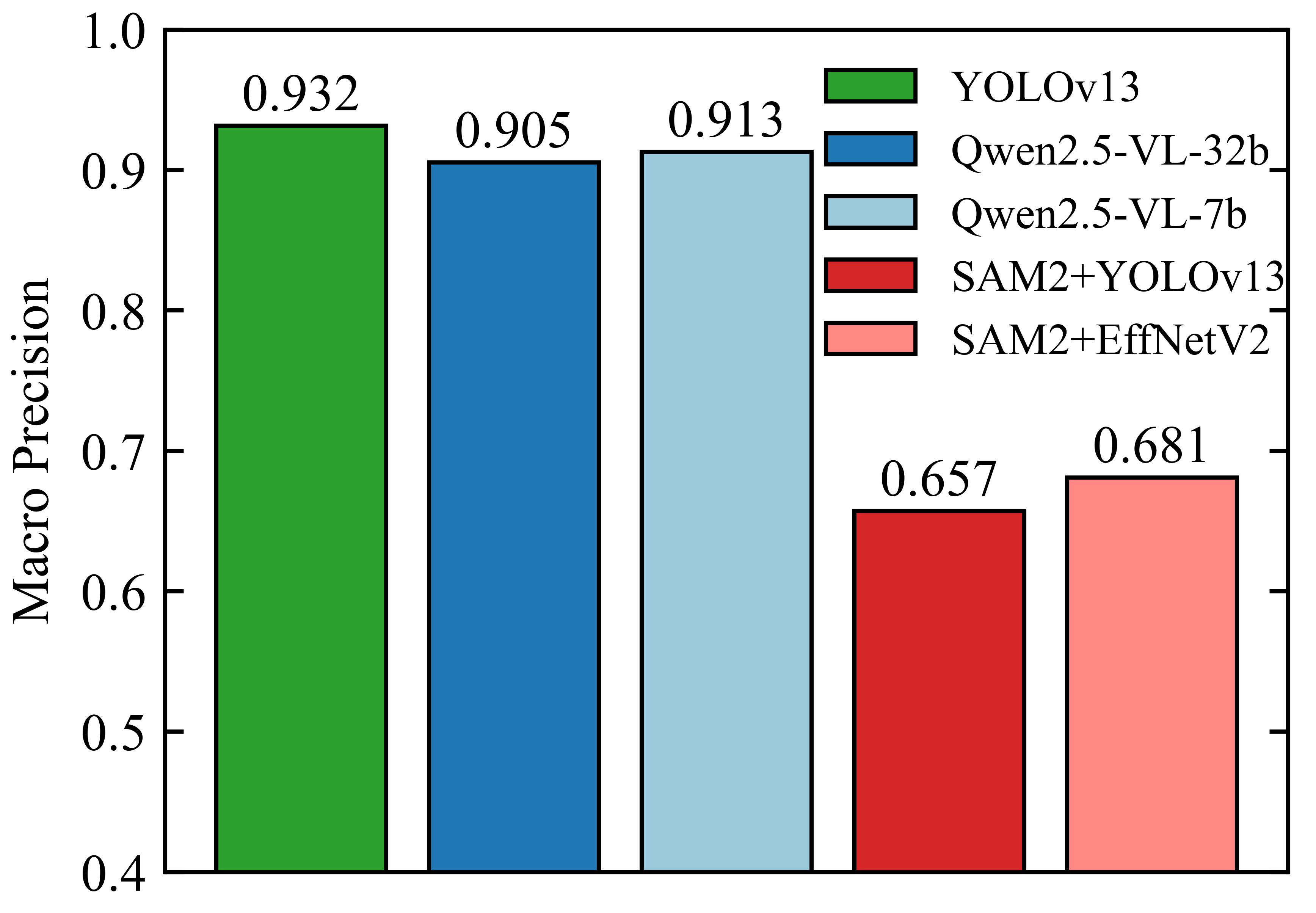}
    \caption{macro-precision}
    \label{Figure-7a}
  \end{subfigure}
  \hfill
  \begin{subfigure}[b]{0.49\linewidth}
    \centering
    \includegraphics[width=\linewidth]{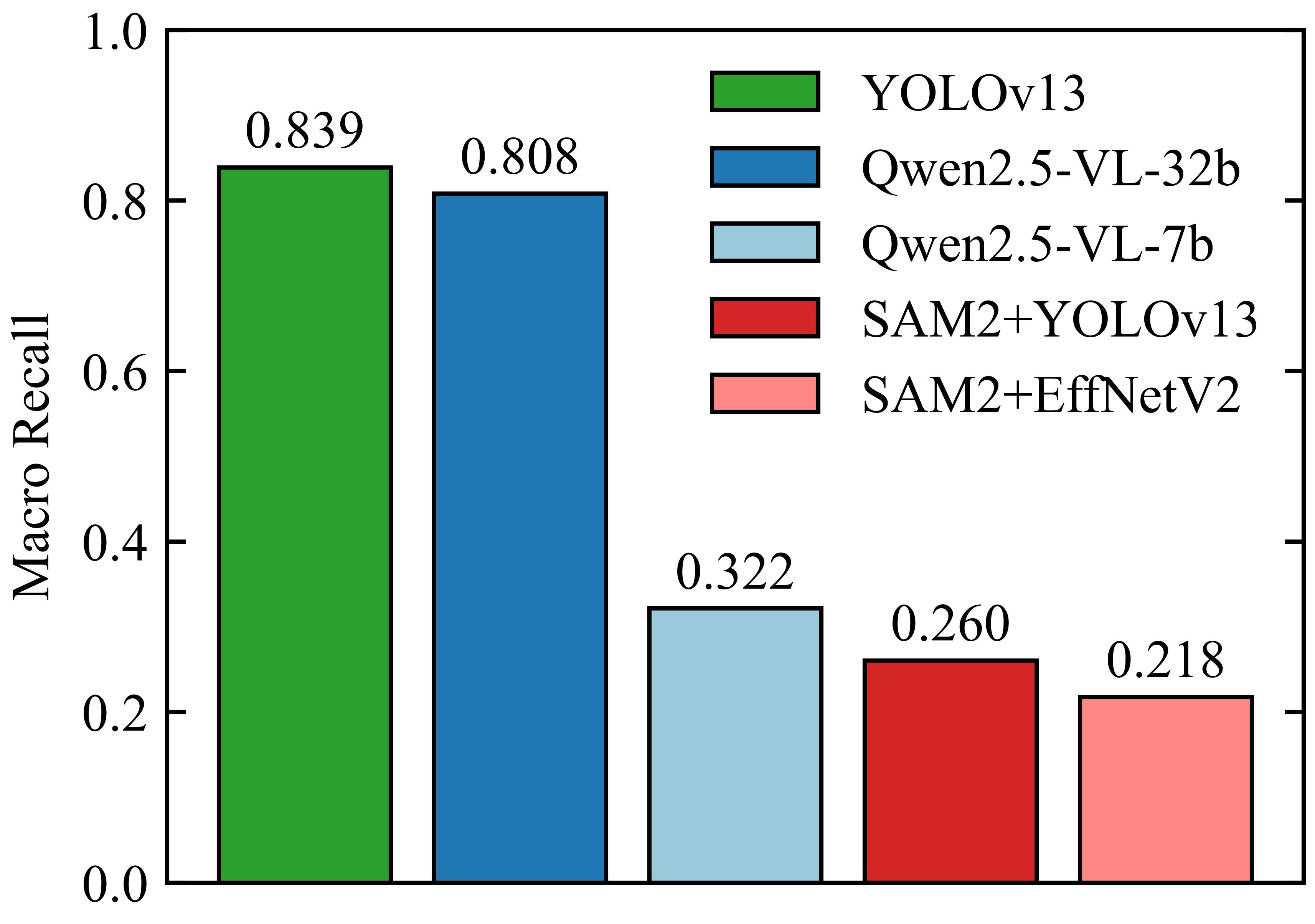}
    \caption{macro-recall}
    \label{Figure-7b}
  \end{subfigure}
  \caption{Performance comparison.}
  \label{Figure-7}
\end{figure}

To gain a deeper understanding of the models' behavior, their performance was further assessed using macro-precision and macro-recall and the results are shown in Figure~\ref{Figure-7}. Macro-precision quantifies the proportion of predicted positive instances that are correct. The analysis in Figure~\ref{Figure-7a} reveals a crucial observation: the top three methods (YOLOv13 at 0.932, Qwen2.5-VL-32b at 0.905, and Qwen2.5-VL-7b at 0.913) all achieve very high macro-precision values above 0.90. This indicates that, across classes, when these models assign an object label, the prediction is highly likely to correspond to a true positive. In contrast, macro-recall measures the proportion of ground truth positive instances that are successfully identified, capturing a model's ability to comprehensively detect objects present in the scene. As shown in Figure~\ref{Figure-7b}, a stark disparity in macro-recall is observed. While YOLOv13 (0.839) and Qwen2.5-VL-32b (0.808) maintain strong macro-recall, the other three methods exhibit very low macro-recall values, with the two SAM2-based pipelines achieving macro-recall values of only 0.260 and 0.218. This suggests that, despite producing relatively accurate predictions when detections occur, these methods fail to identify a substantial fraction of ground-truth objects.

\subsection{Performance Analysis Across Scenarios}\label{performance-analysis-across-scenarios}

\begin{figure}[htbp]
  \centering
  \begin{adjustbox}{max width=\linewidth}
    \includegraphics{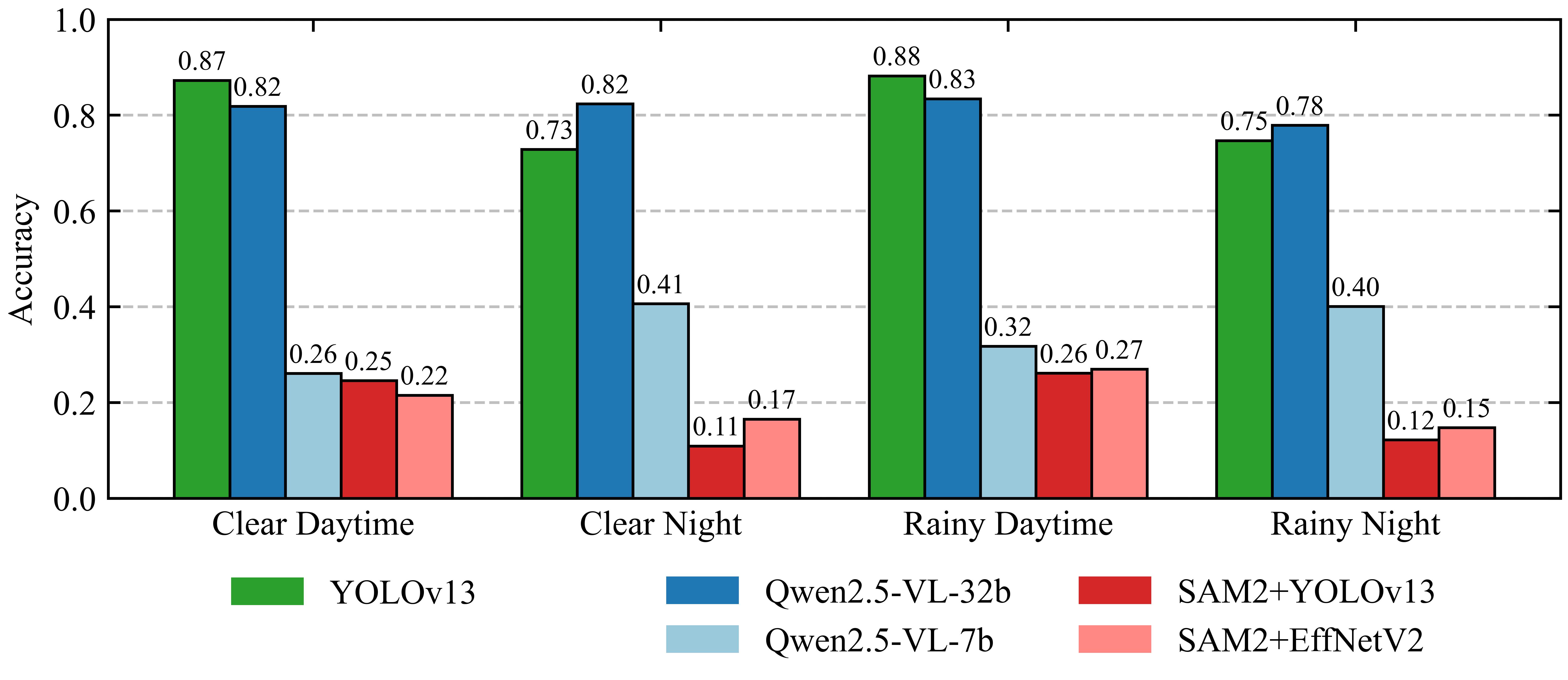}
  \end{adjustbox}
  \caption{Comparison of accuracies across four environmental scenarios.}
  \label{Figure-8}
\end{figure}

To evaluate the robustness of these methods, their performance was analyzed across the four environmental scenarios. The overall accuracies are shown in Figure~\ref{Figure-8}. A primary observation is that for nearly all methods, the highest performance is achieved under the Clear Daytime condition, while the Rainy Night condition proves to be the most challenging scenario. Across all four conditions, the YOLOv13 and Qwen2.5-VL-32b models demonstrate significantly better performance over the other methods. This accuracy comparison reveals a noteworthy pattern. Although YOLOv13 achieved a slightly higher accuracy in Clear Daytime (0.87 vs. 0.82), the Qwen2.5-VL-32b model is the top performer across the remaining three more challenging scenarios, particularly in nighttime settings (Clear Night: 0.82 vs. 0.73; Rainy Night: 0.78 vs. 0.75). This observation suggests that the VLM exhibits greater resilience under low-light conditions, warranting a deeper examination using more robust evaluation metrics.

\begin{figure}[htbp]
  \centering
  \begin{adjustbox}{max width=\linewidth}
    \includegraphics{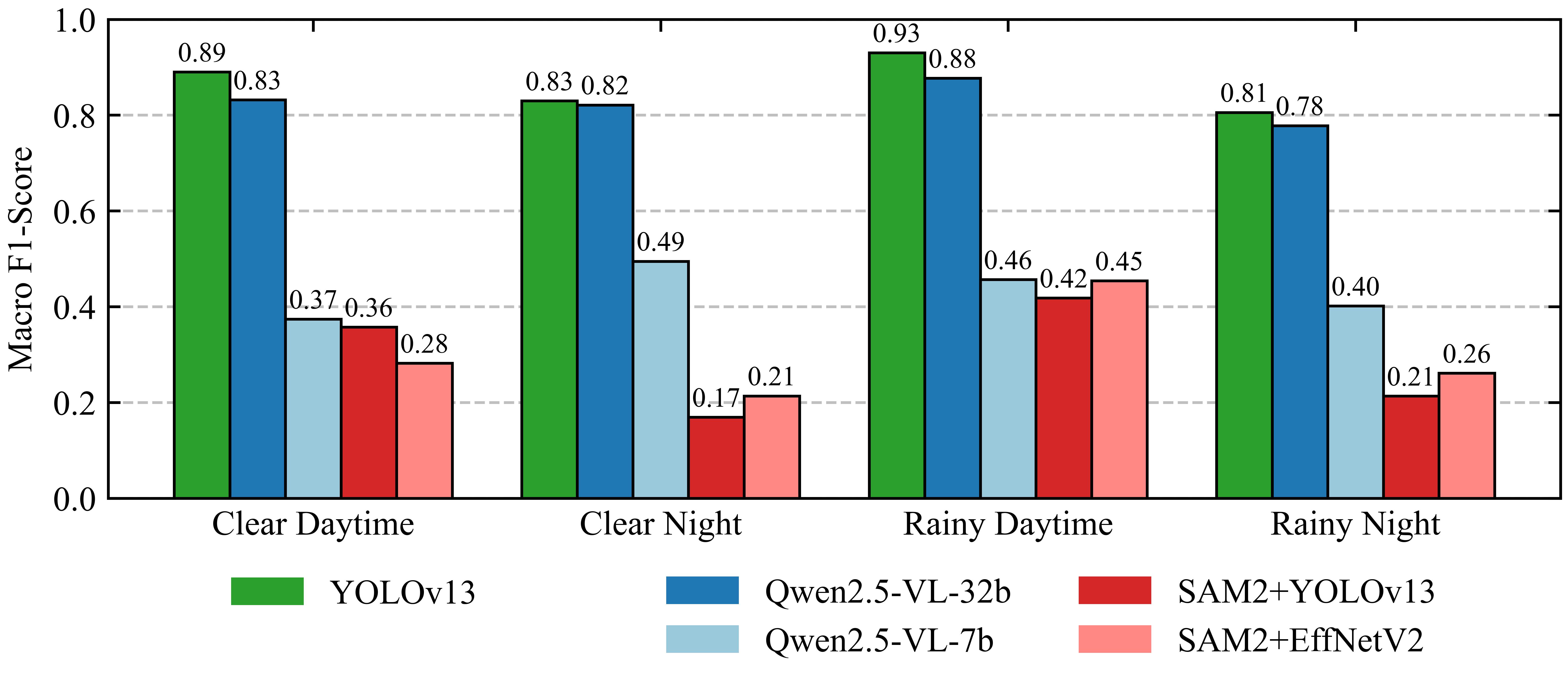}
  \end{adjustbox}
  \caption{Comparison of macro F1-score across four environmental scenarios.}
  \label{Figure-9}
\end{figure}

A more rigorous analysis of model robustness was conducted using the macro F1-score, which accounts for class imbalance. As shown in Figure~\ref{Figure-9}, the results corroborate the overall performance ranking and degradation trends observed with accuracy. However, a closer examination reveals that the YOLOv13consistently achieves the highest Macro F1-Score across all four scenarios, underscoring its superior overall performance. For example, in the most challenging Rainy Night scenario, YOLOv13 attains a macro F1-score of 0.81, outperforming Qwen2.5-VL-32b, which achieves 0.78.

\begin{figure}[htbp]
  \centering
  \begin{subfigure}[b]{\linewidth}
    \centering
    \begin{adjustbox}{max width=\linewidth}
      \includegraphics{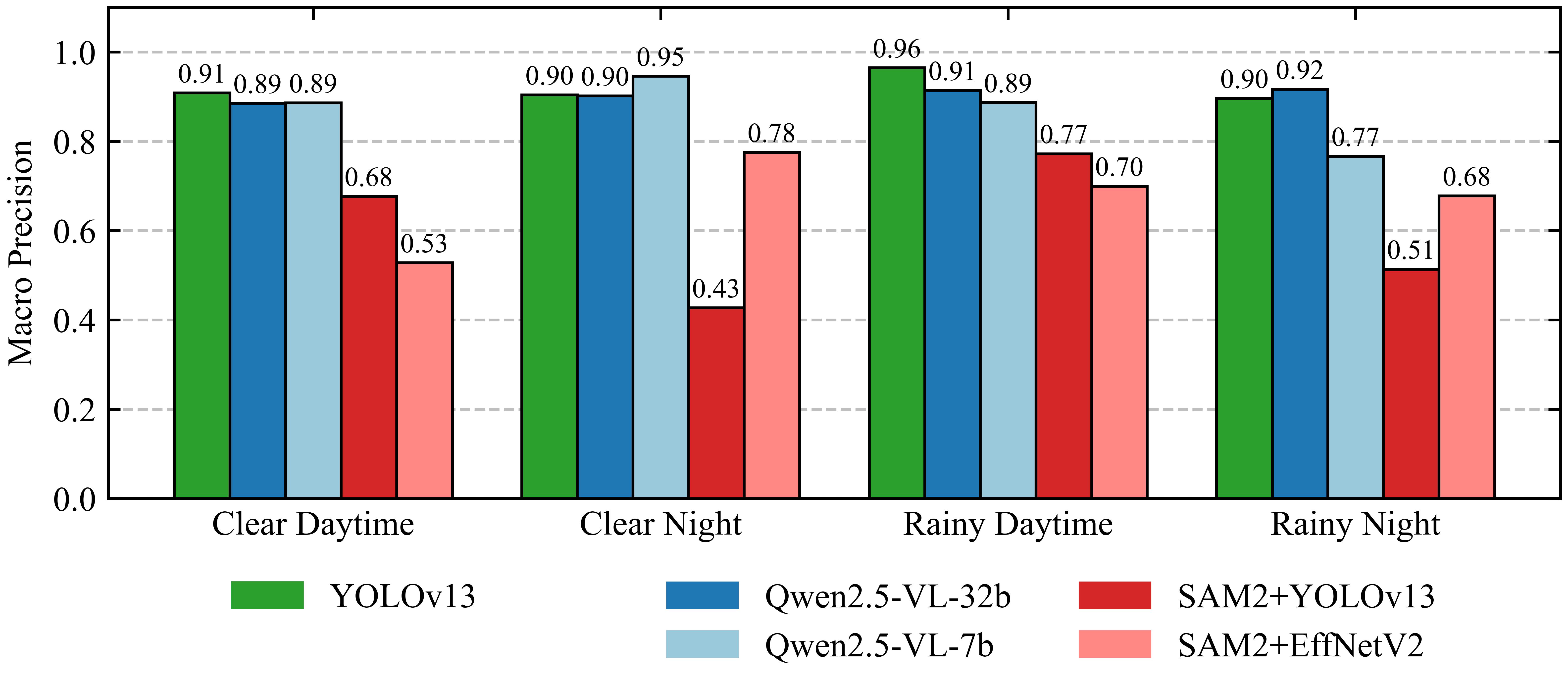}
    \end{adjustbox}
    \caption{macro-precision}
    \label{Figure-10a}
  \end{subfigure}

  \begin{subfigure}[b]{\linewidth}
    \centering
    \begin{adjustbox}{max width=\linewidth}
      \includegraphics{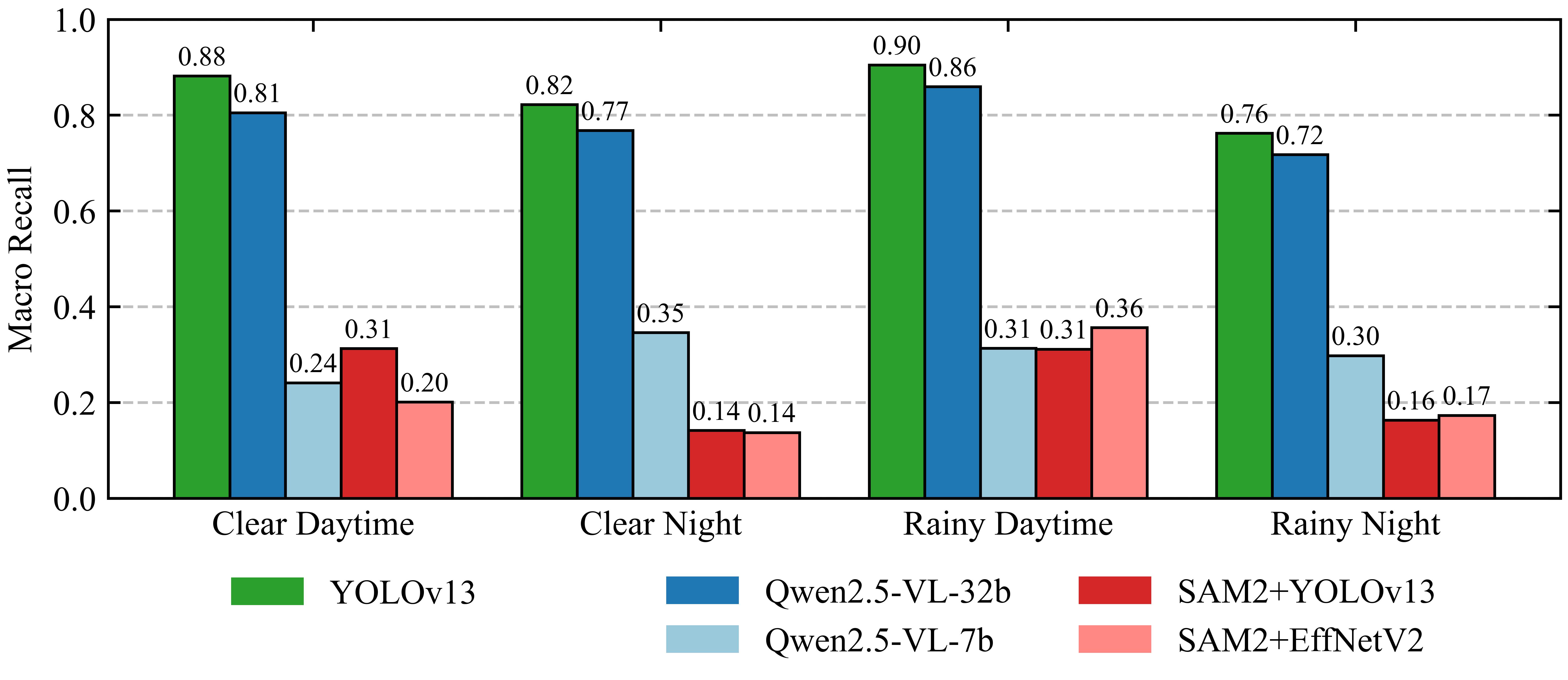}
    \end{adjustbox}
    \caption{macro-recall across four environmental scenarios}
    \label{Figure-10b}
  \end{subfigure}

  \caption{Performance comparison.}
  \label{Figure-10}
\end{figure}

Besides the macro F1-score, Figure~\ref{Figure-10} shows granular performance comparison in macro-precision and macro-recall. Figure~\ref{Figure-10a} shows that macro-precision for the top-performing models (YOLOv13 and Qwen2.5-VL-32B) remains consistently high across all scenarios. In contrast, macro-recall exhibits a clear and systematic decline as environmental conditions worsen (Figure~\ref{Figure-10b}). For example, YOLOv13's recall decreases from 0.88 in Clear Daytime to 0.76 in Rainy Night. This comparison reveals that performance degradation under adverse conditions is driven primarily by missed detections (false negatives), indicating that models increasingly fail to detect objects under challenging visual conditions.

\subsection{Performance Analysis Across Object Categories}\label{performance-analysis-across-object-categories}

Following the analysis of performance across different environmental scenarios, this section examines the performance of each method across five object categories. The goal is to uncover paradigm-specific strengths and limitations with respect to different object types, such as large vehicles versus small signals and common versus rare classes. Similarly, the analysis begins with a general overall accuracy, followed by a more rigorous assessment using the F1-score, precision, and recall to provide a balanced and comprehensive evaluation.

\begin{figure}[htbp]
  \centering
  \begin{adjustbox}{max width=\linewidth}
    \includegraphics{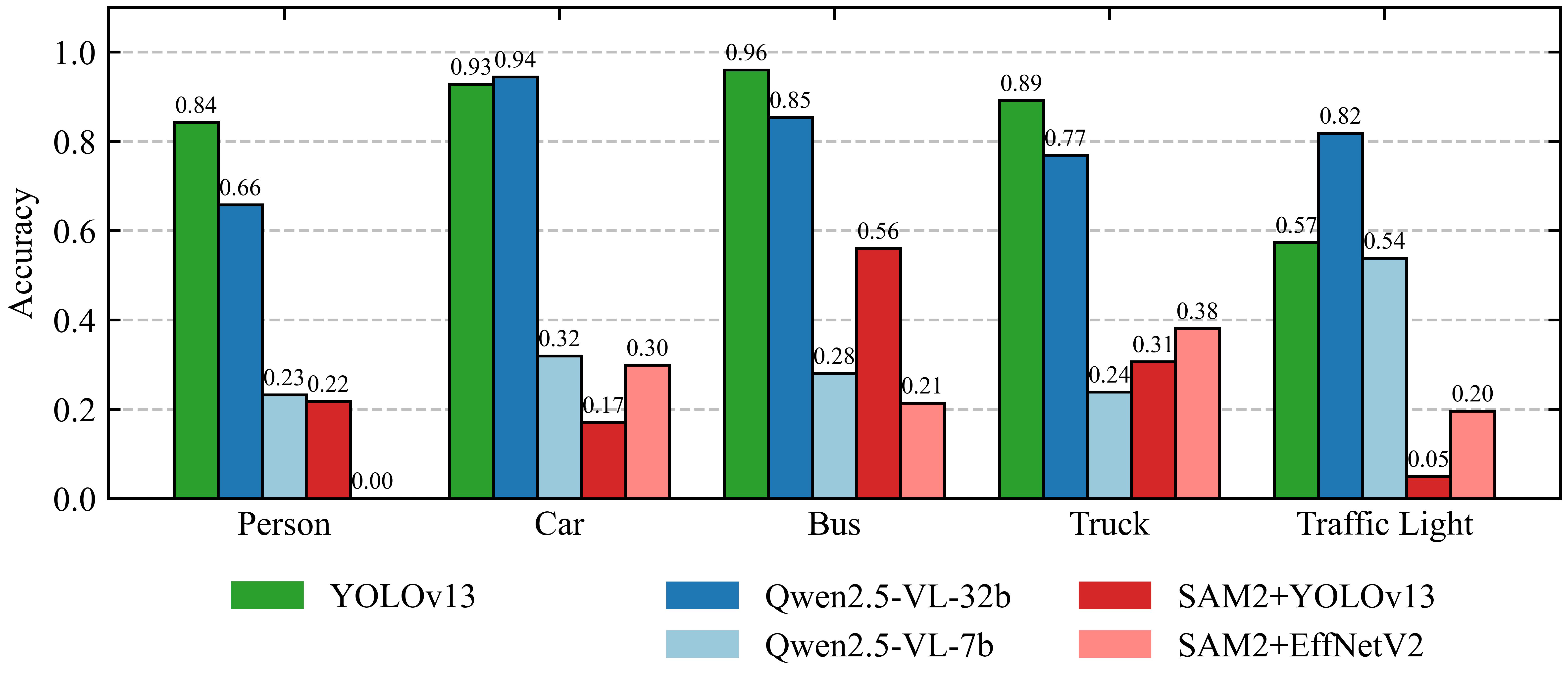}
  \end{adjustbox}
  \caption{Method performance (Accuracy) across five object categories.}
  \label{Figure-11}
\end{figure}

Figure~\ref{Figure-11} presents overall accuracies across the five object categories. Again, YOLOv13 and Qwen2.5-VL-32b emerge as the top performers in most categories, while revealing complementary strengths. YOLOv13 excels on vehicle classes such as Car (0.93) and Bus (0.96), whereas Qwen2.5-VL-32b performs notably better on Traffic Light detection, achieving a much higher accuracy of 0.82 compared to YOLOv13's 0.57.

\begin{figure}[htbp]
  \centering
  \begin{adjustbox}{max width=\linewidth}
    \includegraphics{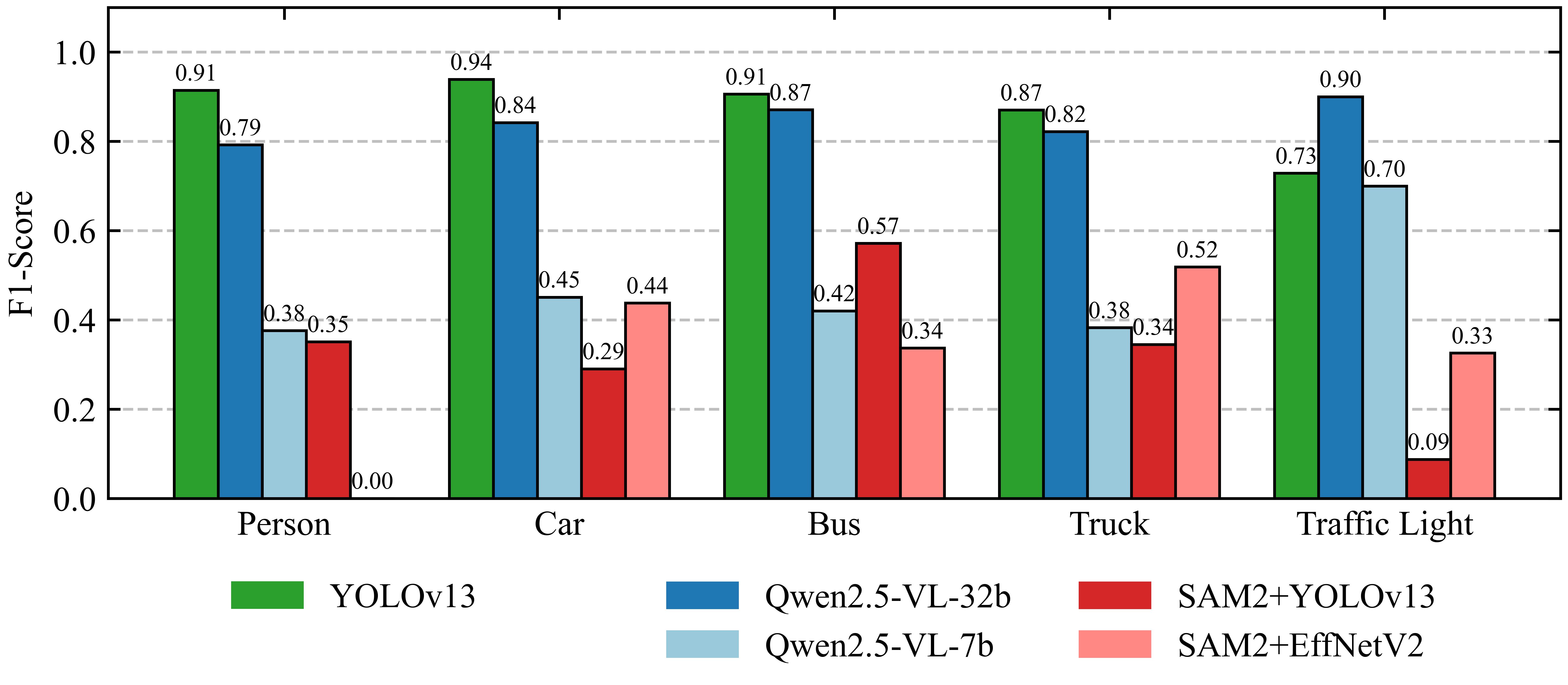}
  \end{adjustbox}
  \caption{Method performance (F1-Score) across five object categories.}
  \label{Figure-12}
\end{figure}

Figure~\ref{Figure-12} shows F1-score across the five object categories, confirming the overall dominance of the YOLOv13 and Qwen2.5-VL-32b. YOLOv13 achieves the highest F1-scores on common, large objects: Person (0.91), Car (0.94), Bus (0.91), and Truck (0.87), highlighting its strength in recognizing well-defined classes. Notably, a clear performance crossover emerges for the Traffic Light category: Qwen2.5-VL-32B attains a substantially higher F1-score (0.90) than YOLOv13 (0.73), revealing a distinct advantage of the large VLM paradigm for small and visually challenging objects.

\begin{figure}[htbp]
  \centering
  \begin{subfigure}[b]{\linewidth}
    \centering
    \begin{adjustbox}{max width=\linewidth}
      \includegraphics{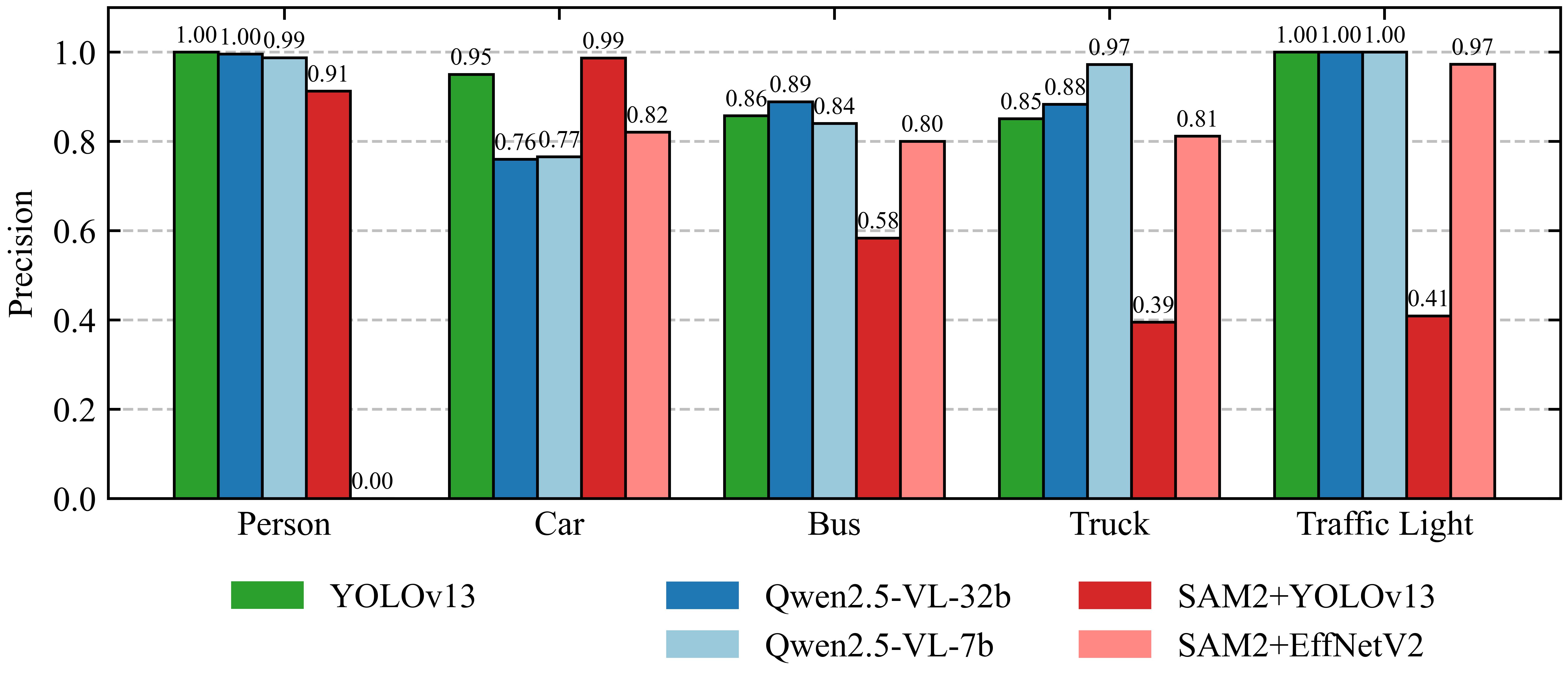}
    \end{adjustbox}
    \caption{precision}
    \label{Figure-13a}
  \end{subfigure}

  \begin{subfigure}[b]{\linewidth}
    \centering
    \begin{adjustbox}{max width=\linewidth}
      \includegraphics{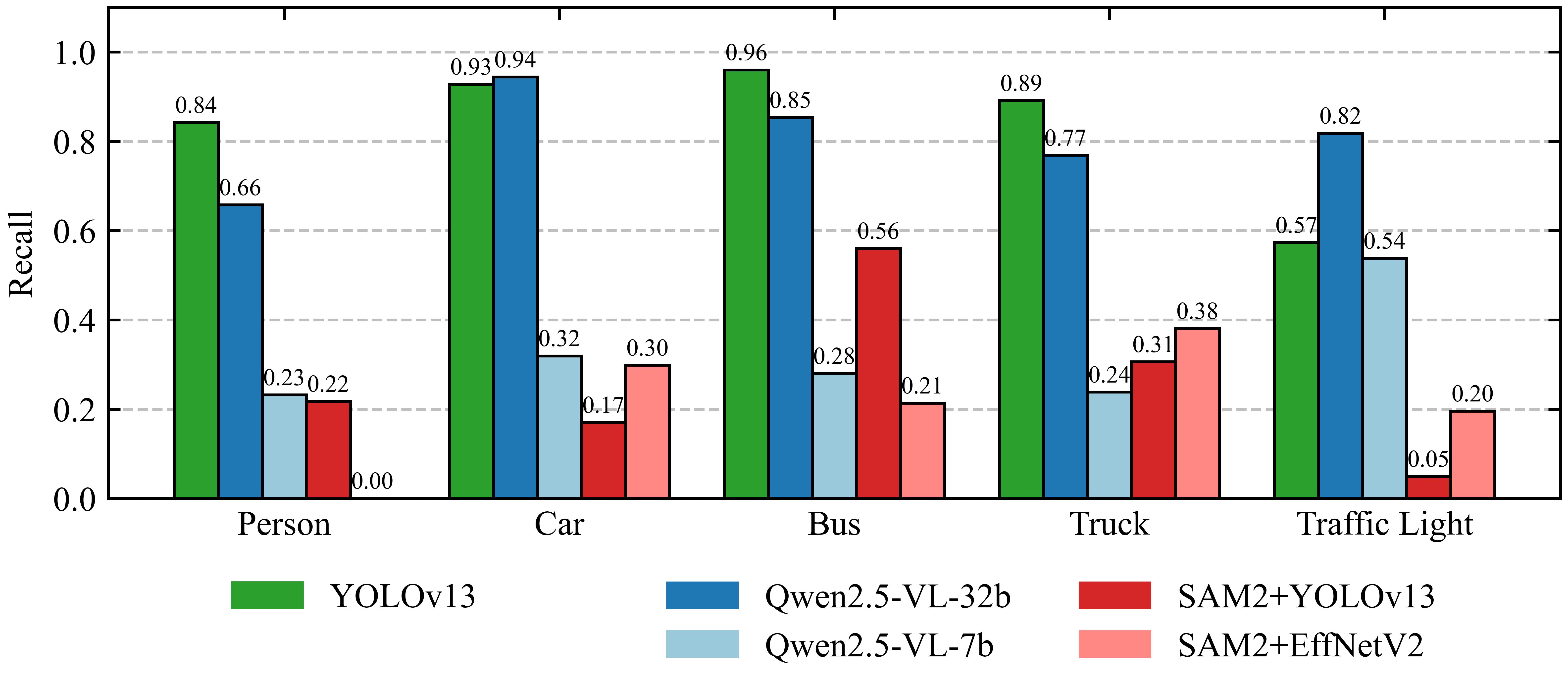}
    \end{adjustbox}
    \caption{recall across object categories}
    \label{Figure-13b}
  \end{subfigure}

  \caption{Performance comparison.}
  \label{Figure-13}
\end{figure}

Figure~\ref{Figure-13}. shows class-level precision and recall. Both YOLOv13 and Qwen2.5-VL-32B achieve perfect precision (1.00) for the Traffic Light class, indicating no false positives. The performance gap is instead driven by recall: Qwen2.5-VL-32B attains a recall of 0.82, substantially higher than YOLOv13's 0.57. This explains YOLOv13's lower F1-score, which stems from missed detections rather than misclassification. The VLM's superior recall likely reflects its ability to exploit global scene context and implicit world knowledge, enabling more reliable detection of small, visually challenging objects.

\subsection{Deeper Analysis of Model Behavior}\label{deeper-analysis-of-model-behavior}

To complement the quantitative results, this section provides a deeper analysis of the internal behavior and qualitative error patterns of the evaluated models. This analysis proceeds in two parts: first, confusion matrices are used to visualize the class-specific error modes; second, confidence score distributions of key model components are examined to understand how the models arrive at their decisions and to diagnose the underlying causes of failures.

\begin{figure}[htbp]
  \centering
  \begin{subfigure}[b]{0.49\linewidth}
    \centering
    \includegraphics[width=\linewidth]{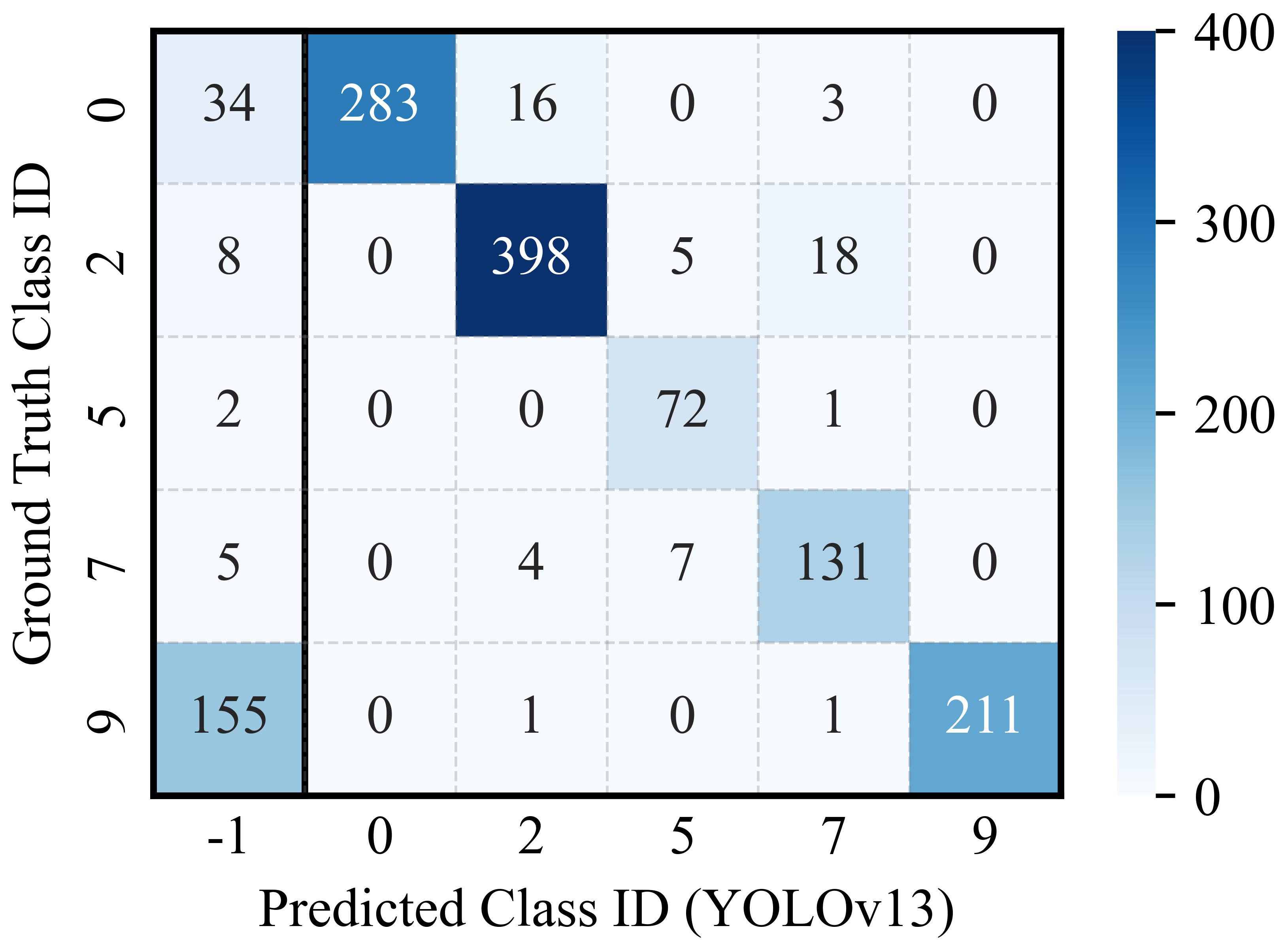}
    \caption{YOLOv13}
    \label{Figure-14a}
  \end{subfigure}
  \hfill
  \begin{subfigure}[b]{0.49\linewidth}
    \centering
    \includegraphics[width=\linewidth]{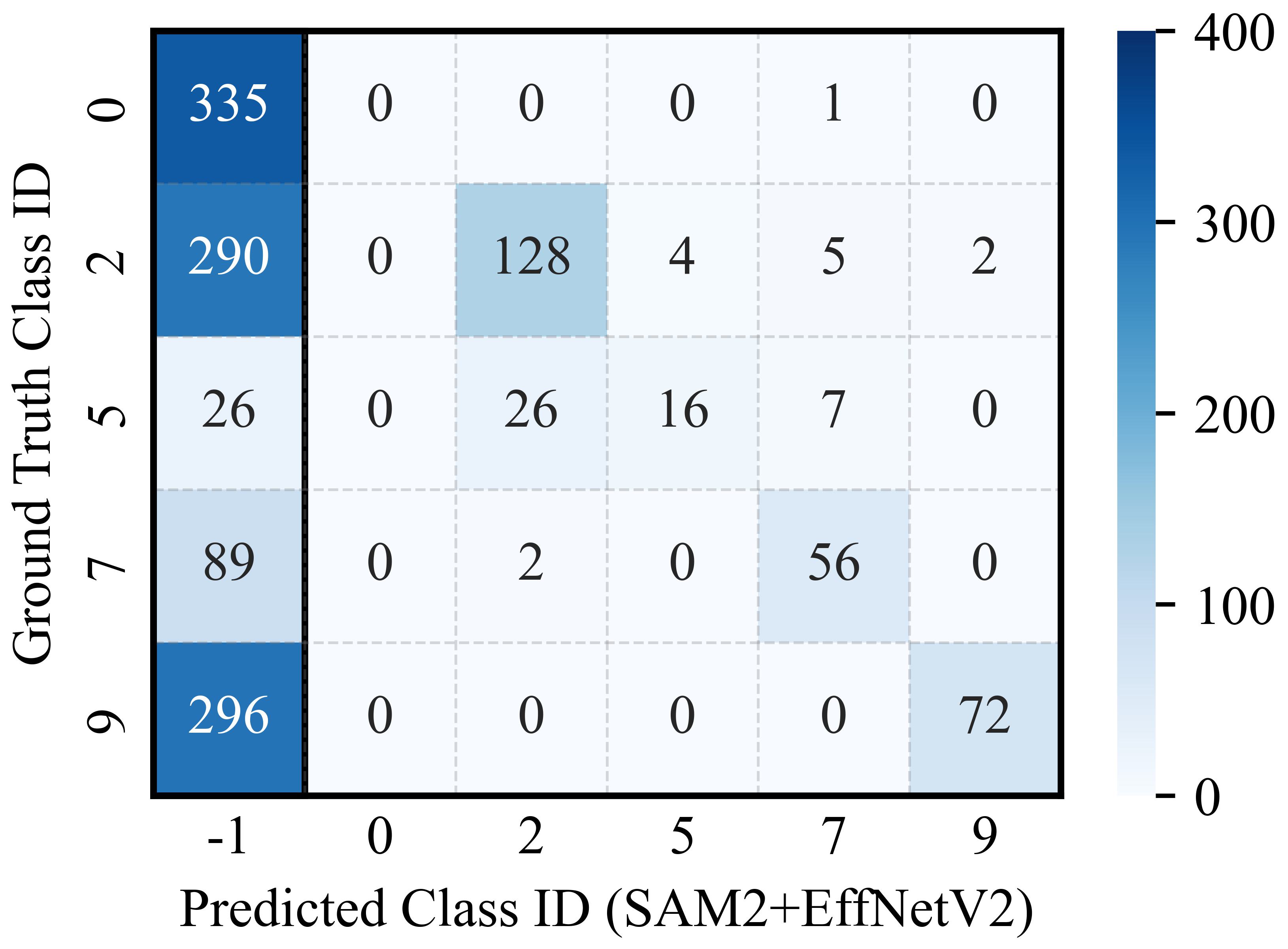}
    \caption{SAM2+EfficientNetV2}
    \label{Figure-14b}
  \end{subfigure}

  \vspace{1em}

  \begin{subfigure}[b]{0.49\linewidth}
    \centering
    \includegraphics[width=\linewidth]{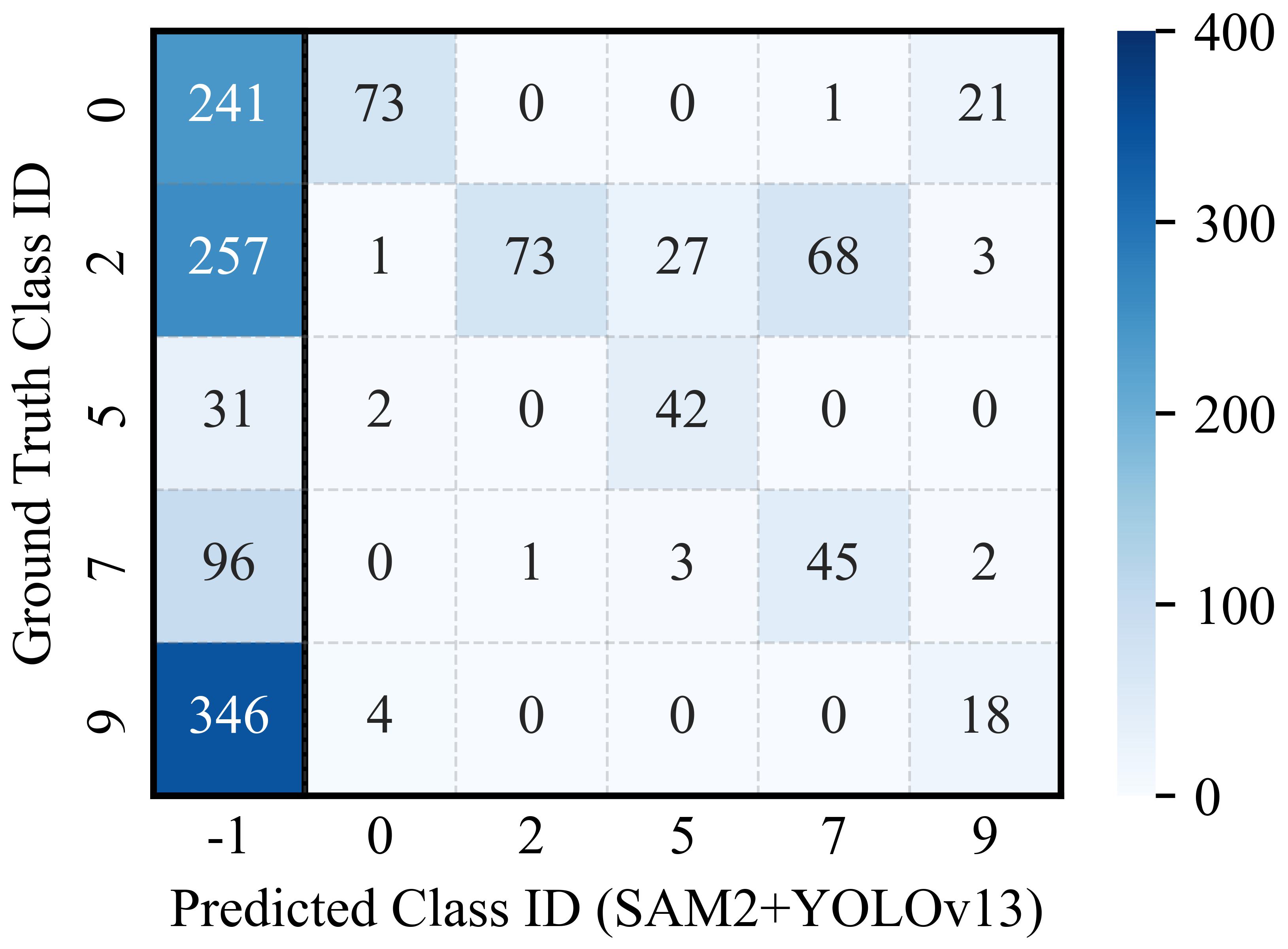}
    \caption{SAM2+YOLOv13}
    \label{Figure-14c}
  \end{subfigure}
  \hfill
  \begin{subfigure}[b]{0.49\linewidth}
    \centering
    \includegraphics[width=\linewidth]{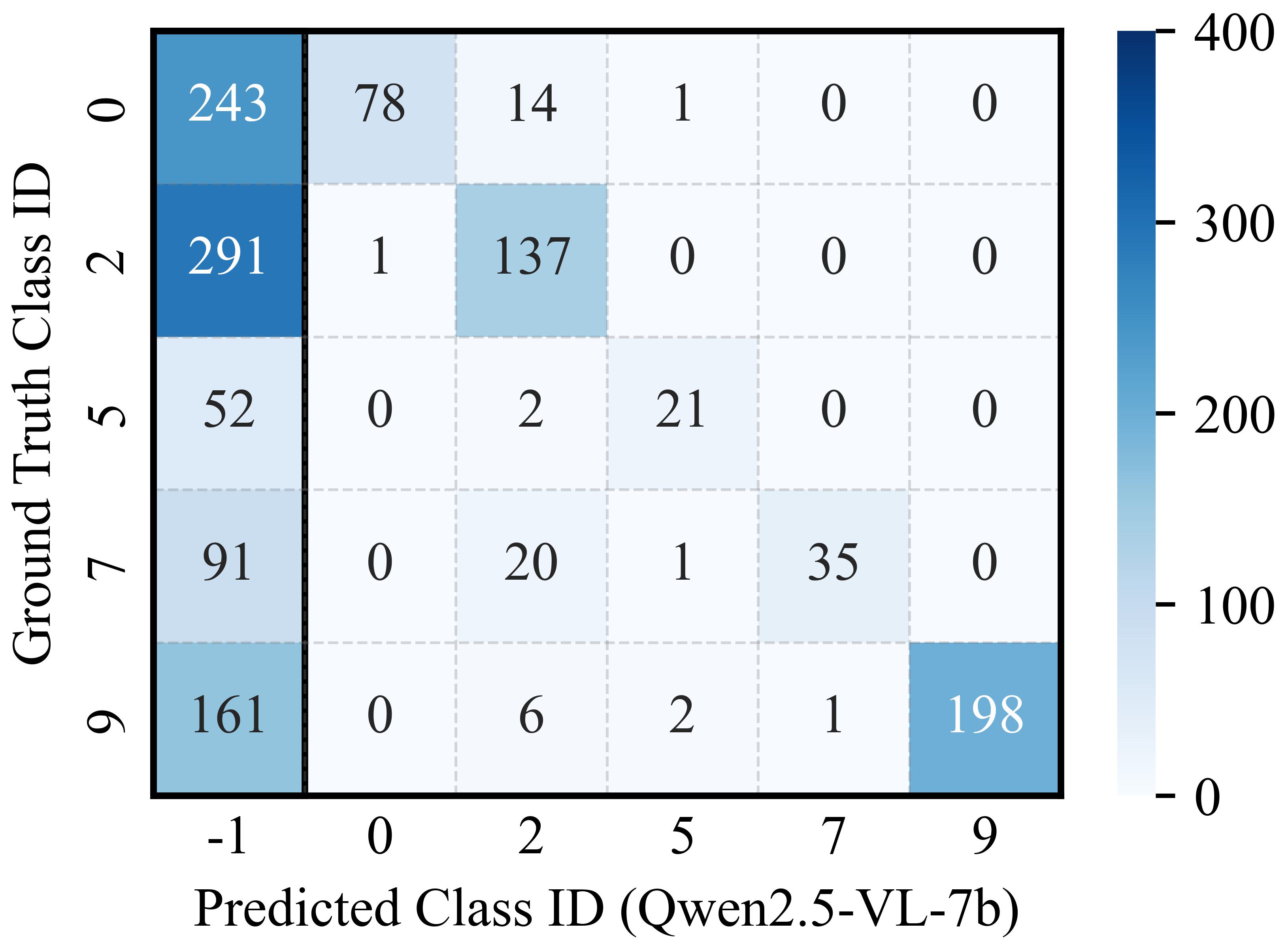}
    \caption{Qwen2.5-VL-7b}
    \label{Figure-14d}
  \end{subfigure}

  \vspace{1em}

  \begin{subfigure}[b]{0.49\linewidth}
    \centering
    \includegraphics[width=\linewidth]{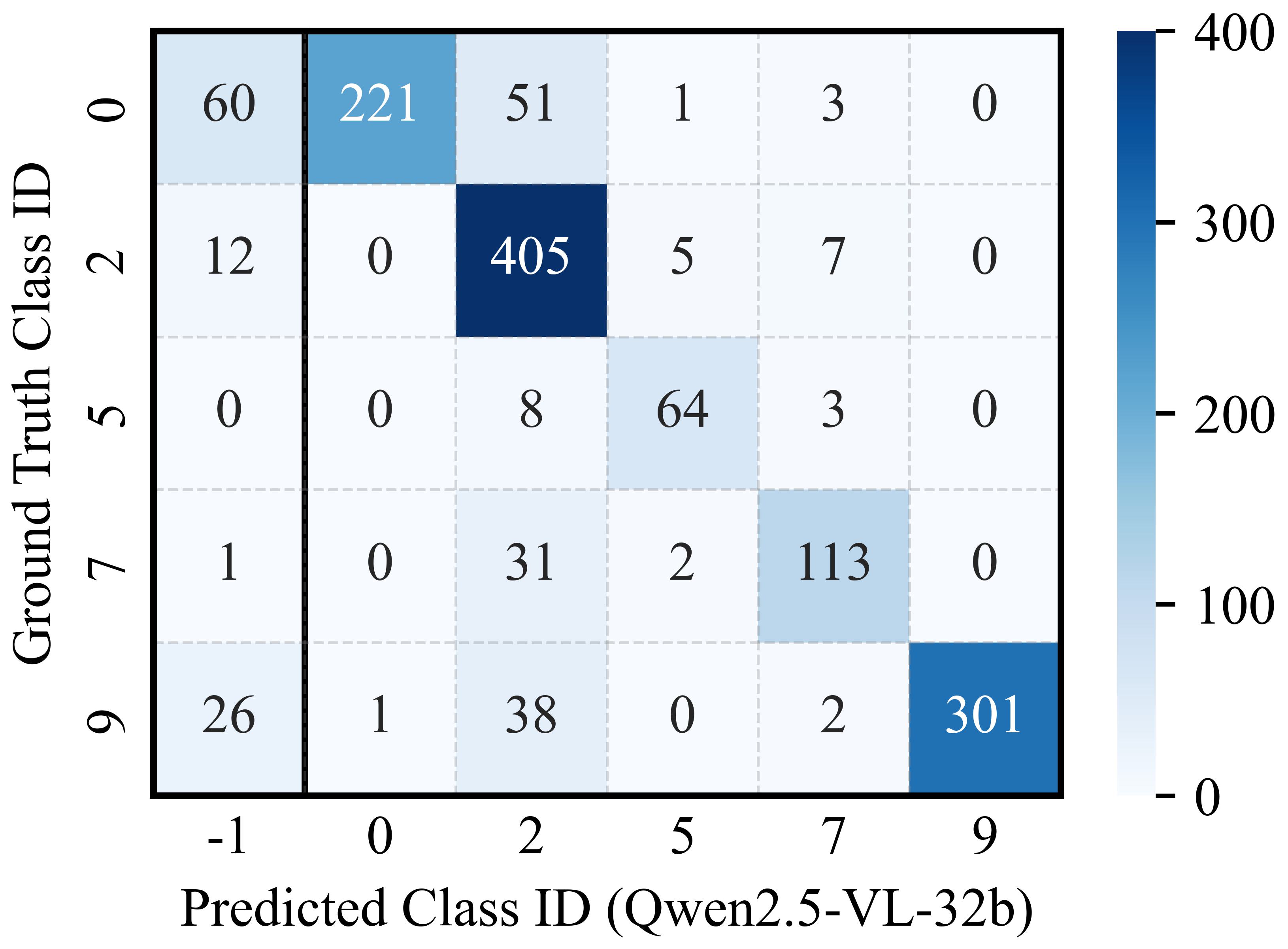}
    \caption{Qwen2.5-VL-32b}
    \label{Figure-14e}
  \end{subfigure}

  \caption{Confusion matrices for all five methods.}
  \label{Figure-14}
\end{figure}

Figure~\ref{Figure-14} presents the confusion matrices for all five evaluated methods, allowing for a direct comparison of their error patterns. The matrices for the top-performing models, YOLOv13 (Figure~\ref{Figure-14a}) and Qwen2.5-VL-32b (Figure~\ref{Figure-14e}), exhibit strong diagonal dominance, visually confirming their high rate of correct classifications. For these models, the few off-diagonal errors represent logical misclassifications, such as the confusion between Truck (ID 7) and Car (ID 2), which is expected given their visual similarity. The most critical insight, however, comes from the first column of each matrix, which represents predictions of class ID -1. The matrices for both segmentation-assisted methods, SAM2+EfficientNetV2 (Figure~\ref{Figure-14b}) and SAM2+YOLOv13 (Figure~\ref{Figure-14c}), show extremely high values in this column. For instance, in Figure~\ref{Figure-14b}, 335 out of 336 Person instances were incorrectly rejected. Similarly, the matrix for the smaller VLM, Qwen2.5-VL-7b (Figure~\ref{Figure-14d}), also shows a very high number of rejections in this column compared to its larger counterpart. This provides direct visual evidence for the previously established finding that the catastrophic failure of the segmentation-assisted methods, and the comparatively weaker performance of the 7b VLM, is primarily due to an inability to make a positive classification (a recall problem), not a high rate of misclassification.

\begin{figure}[htbp]
  \centering
  \begin{subfigure}[b]{0.49\linewidth}
    \centering
    \includegraphics[width=\linewidth]{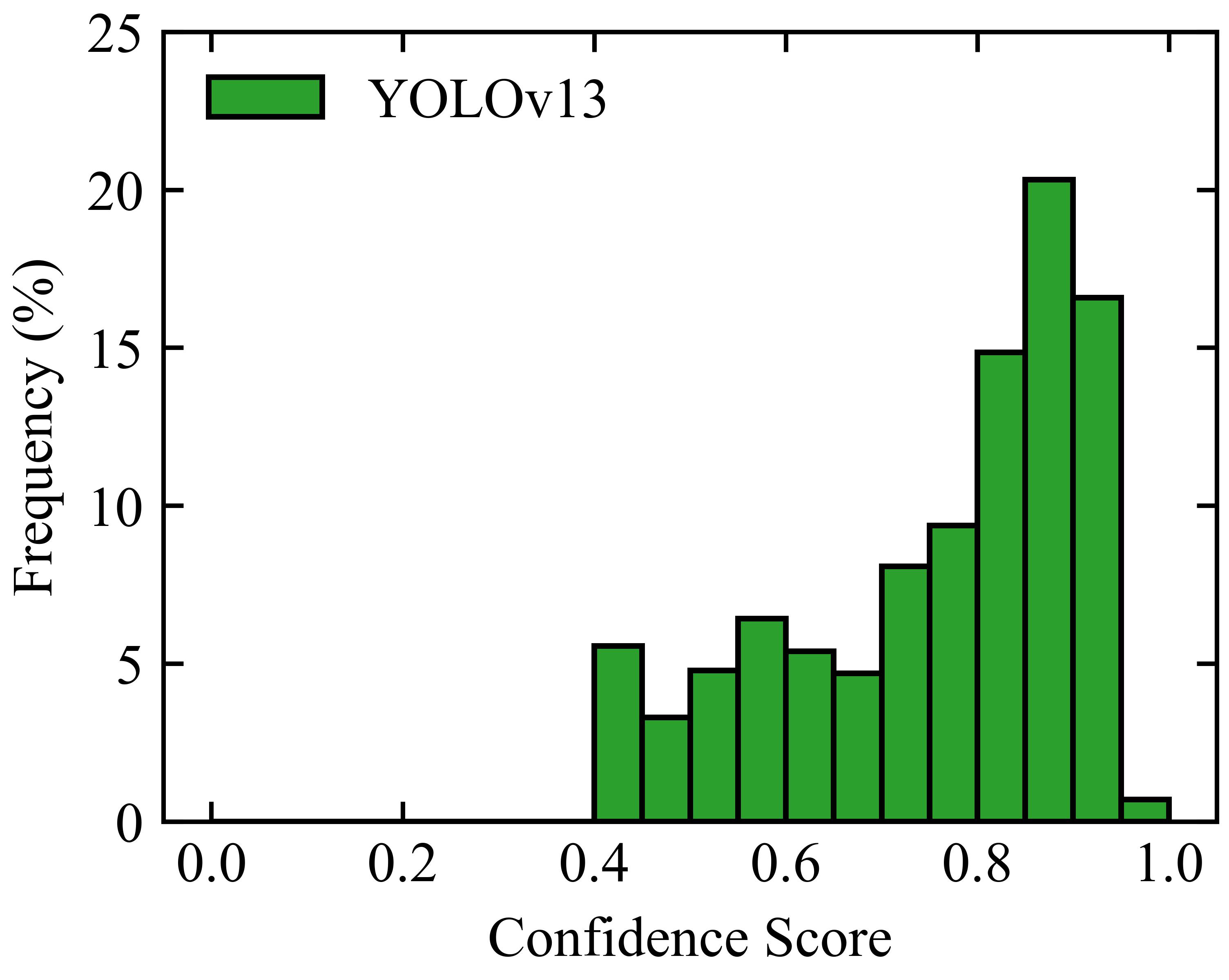}
    \caption{YOLOv13}
    \label{Figure-15a}
  \end{subfigure}
  \hfill
  \begin{subfigure}[b]{0.49\linewidth}
    \centering
    \includegraphics[width=\linewidth]{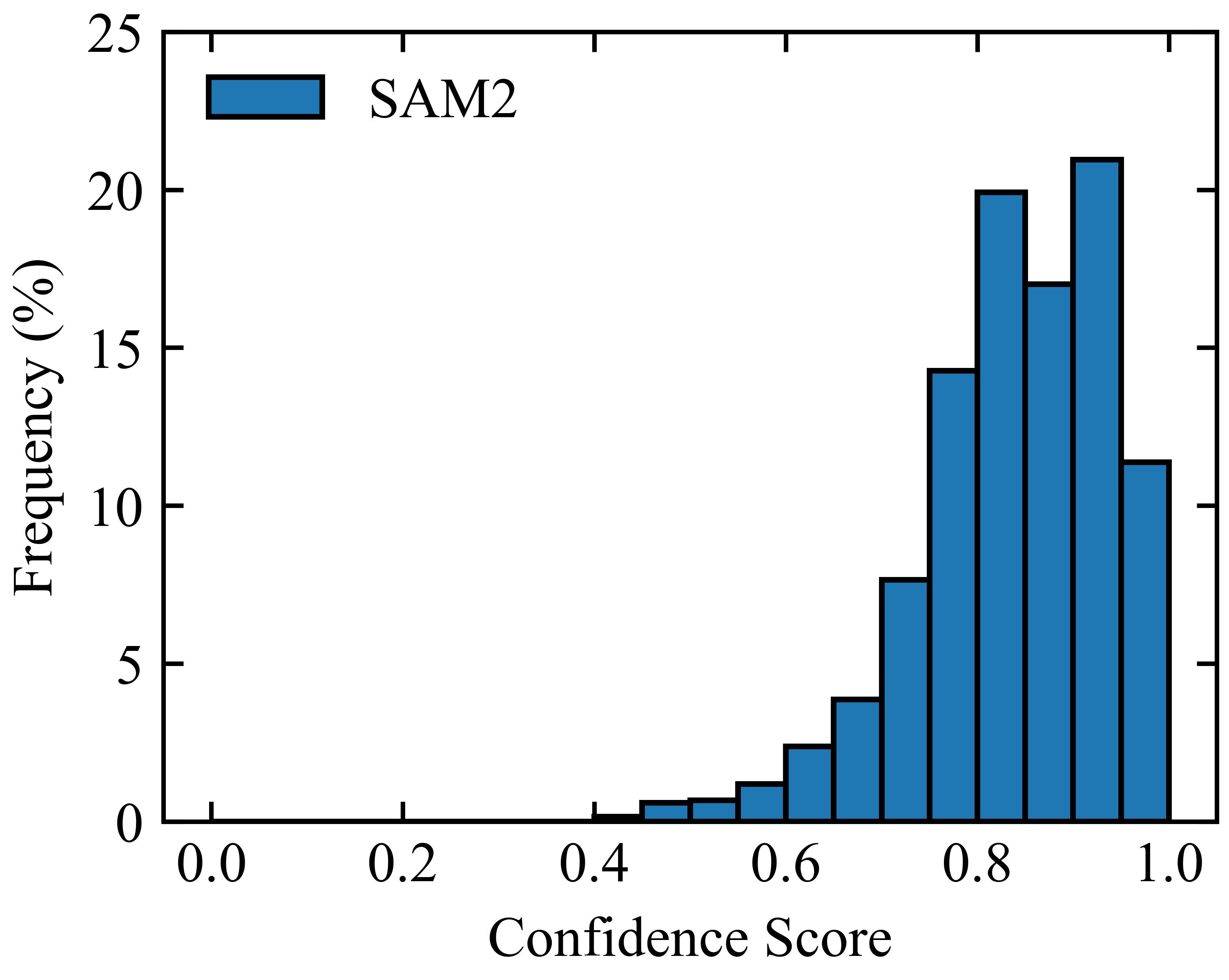}
    \caption{SAM2}
    \label{Figure-15b}
  \end{subfigure}

  \vspace{1em} 

  \begin{subfigure}[b]{0.49\linewidth}
    \centering
    \includegraphics[width=\linewidth]{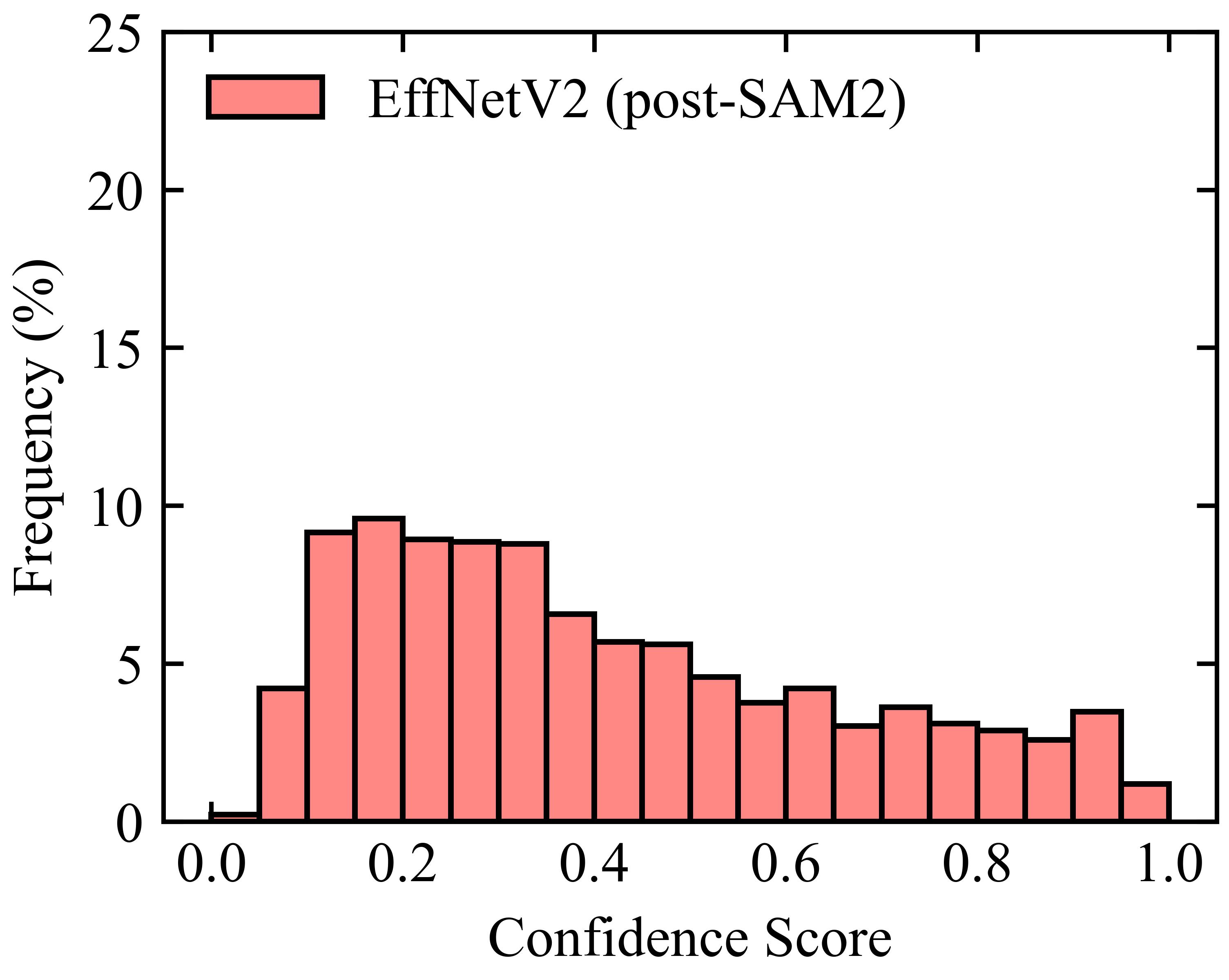}
    \caption{EfficientNetV2}
    \label{Figure-15c}
  \end{subfigure}
  \hfill
  \begin{subfigure}[b]{0.49\linewidth}
    \centering
    \includegraphics[width=\linewidth]{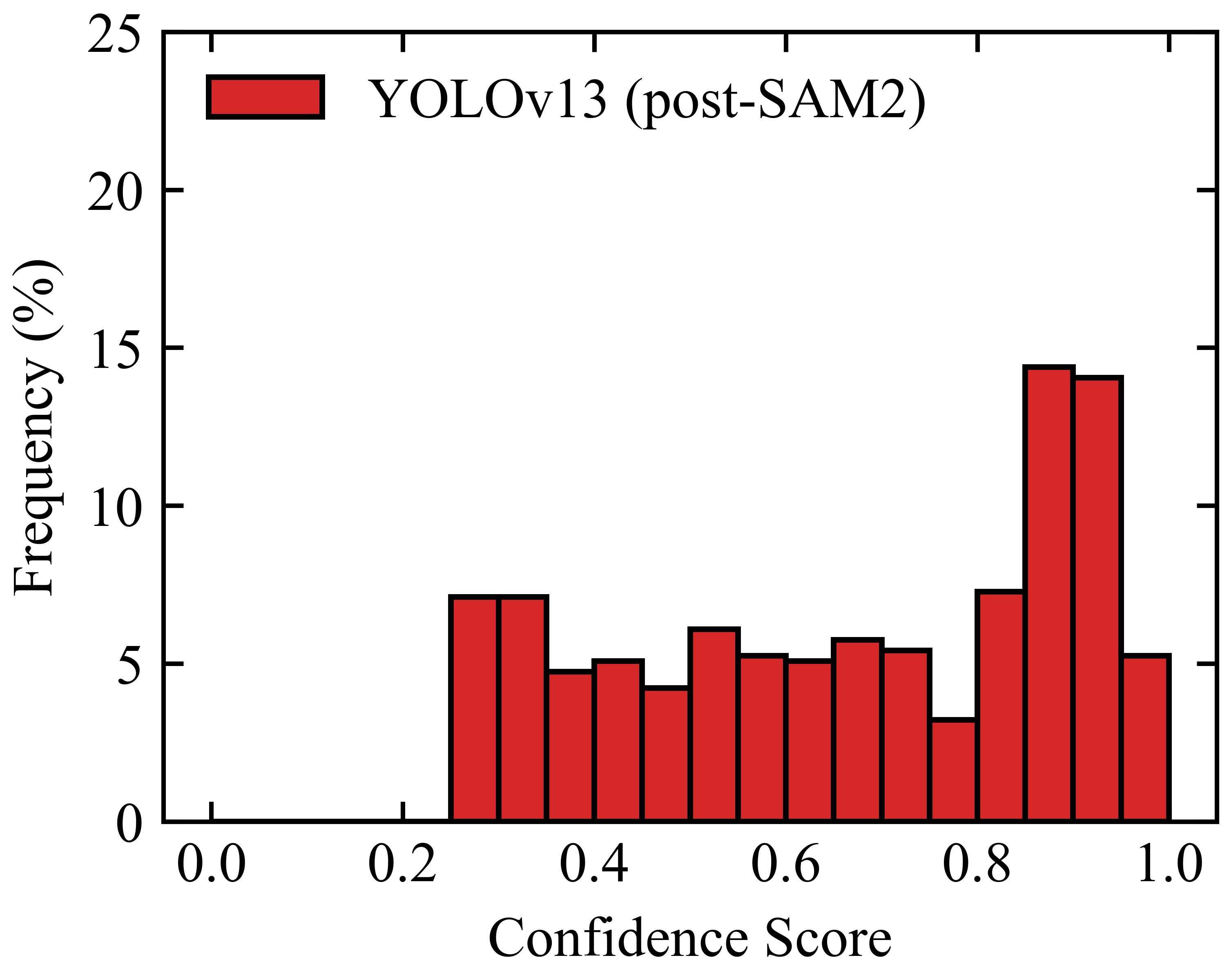}
    \caption{YOLOv13 (post-SAM2)}
    \label{Figure-15d}
  \end{subfigure}

  \caption{Confidence score distributions for model components.}
  \label{Figure-15}
\end{figure}

To further diagnose the underlying cause of the segmentation-assisted paradigm's failure, the confidence scores of its constituent components were analyzed, as shown in Figure~\ref{Figure-15}. The histogram for the standalone YOLOv13 model in Figure~\ref{Figure-15a} shows that it is generally very confident in its predictions, with a distribution heavily skewed towards high scores between 0.8 and 1.0. The key piece of evidence is presented in Figure~\ref{Figure-15b}, which shows that the SAM2 segmentation stage is also very confident, with its scores heavily concentrated above 0.8. However, when the classifiers are run on the crops generated by SAM2, their confidence decreases significantly. As seen in Figure~\ref{Figure-15c} and Figure~\ref{Figure-15d}, the post-SAM2 classifiers exhibit much more uncertain and scattered confidence distributions, with the EfficientNetV2 model's confidence scores concentrated in the low 0.1 to 0.3 range.

These observations lead to a clear diagnosis of the ``part-versus-whole'' problem. The failure of the SAM2-based pipeline does not stem from poor or low-confidence segmentation; rather, SAM2 is often highly confident in its outputs. The issue arises because SAM2 confidently segments an incorrect semantic entity. For instance, when a driver's gaze falls on a car's taillight, SAM2 may accurately segment the taillight as an object part. The downstream classifier, however, is then presented with an isolated part and must select from whole-object categories such as Car, leading to uncertainty or misclassification. This semantic mismatch accounts for the low confidence scores and frequent background (Class ID: -1) predictions, identifying the part-versus-whole problem as a fundamental limitation of the segmentation-assisted paradigm for this task.

\subsection{Computational Cost Analysis}\label{computational-cost-analysis}

Beyond model performance, computational efficiency is a critical for real-world deployment in vehicles. To evaluate this, all experiments were conducted on a single NVIDIA A6000 GPU. Inference speed (FPS) was computed by averaging the processing time over 100 test images, while peak GPU memory usage reflects the maximum memory consumption observed during this process.

Table~\ref{Table-2} summarizes the computational costs and resource requirements of the five methods, revealing stark differences in both speed and memory. YOLOv13 is the only model achieving real-time performance (36.1 FPS) with modest memory use (27.6M parameters, 1.53 GB VRAM), making it suitable for in-vehicle deployment. SAM2-based pipelines run at about 10 FPS (10.3-10.4 FPS), potentially acceptable for near real-time tasks but insufficient for latency-critical applications. The VLMs exhibit extremely low throughput, with Qwen2.5-VL-7B at 2.1 FPS and Qwen2.5-VL-32B at 0.07 FPS (about 14 s per frame), alongside massive memory demands (17.6-45 GB VRAM), necessitating a top-tier or customized GPU.

Combining these results with the previous performance analyses highlights a clear performance-efficiency trade-off. YOLOv13 achieves a strong balance of accuracy and speed, making it the most practical choice for real-time deployment. Qwen2.5-VL-32B, while robust and accurate in certain conditions, incurs prohibitive computational costs, limiting its real-time applicability. SAM2-based methods perform poorly in both efficiency and accuracy, making them the least viable option.

\begin{table}[htbp]
  \centering
  \caption{Computational Cost Comparison of All Methods.}
  \label{Table-2}

  \renewcommand{\tabularxcolumn}[1]{m{#1}}
  
  \begin{tabularx}{\linewidth}{>{\raggedright\arraybackslash}m{2.9cm} >{\centering\arraybackslash}m{2.6cm} *{4}{>{\centering\arraybackslash}X}}
    \toprule
    \textbf{Paradigm} & \textbf{Method} & \textbf{Model Size} & \textbf{Parameters (M)} & \textbf{Inference Speed (FPS)} & \textbf{Peak GPU Memory Usage (GB)} \\
    \midrule
    Object Detection-Based & YOLOv13 & 107 MB & 27.6 & 36.1 & 1.53 GB \\
    \midrule
    \multirow{2}{=}{Segmentation-Assisted Classification} & SAM2 +EfficientNetV2 & 963 MB & 343.0 & 10.4 & 11.83 GB \\
    & SAM2 +YOLOv13 & 1.3 GB & 252.1 & 10.3 & 9.20 GB \\
    \midrule
    \multirow{2}{=}{Vision-Language Model-Based} & Qwen2.5-VL-7b & 15.5 GB & 8,292.2 & 2.1 & 17.63 GB \\
    & Qwen2.5-VL-32b & 63.6 GB & 33,452.7 & 0.07 & 44.97 GB \\
    \bottomrule
  \end{tabularx}
\end{table}

\section{Discussion}\label{discussion}

The principal finding of this study is that direct object detection and large-scale VLM paradigms outperform the segmentation-assisted classification approach for point-based object identification. This suggests that methods leveraging holistic scene context are more effective than those relying on an analytic, staged strategy. Both YOLOv13 and Qwen2.5-VL-32B, despite their vastly different architectures, analyze the entire image to inform their predictions, enabling more robust inferences than the two-stage process of isolating a small, context-free region before classification.

The segmentation-assisted methods fail due to a fundamental ``part-versus-whole'' semantic gap. Confidence analysis (Figure~\ref{Figure-15}) shows that SAM2 produces highly confident segmentations, while downstream classifiers exhibit low and unstable confidence, indicating that errors arise from confidently segmenting object parts rather than complete objects. For vehicles, gaze points often yield precise segmentations of components such as license plates, which are not classifiable as Car. For small objects (Person, Traffic Light), the segmented regions are frequently too small and context-poor for reliable recognition. Performance on Person is further limited by a taxonomy mismatch in ImageNet-1K. In contrast, moderate success on larger objects (Bus, Truck) occurs when gaze points fall on broad object regions, allowing full-object segmentation and more reliable classification.

By contrast, the strong performance of the large VLM, particularly in challenging situations, highlights the advantage of its contextual and commonsense reasoning capabilities. The superiority of Qwen2.5-VL-32b was most evident in its robust performance for small objects like Traffic Light (Figure~\ref{Figure-12}) and its relative stability in adverse conditions (Figure~\ref{Figure-9}). The precision-recall analysis for the Traffic Light class (Figure~\ref{Figure-13}) is especially informative: despite perfect precision, the VLM's substantially higher recall indicates its ability to leverage implicit world knowledge, effectively inferring that a small, colored light at an intersection is a traffic light. This advantage is most pronounced at night, where Qwen2.5-VL-32b significantly outperforms the traditional detector. Such holistic scene reasoning enables the VLM to remain effective even when local visual cues are degraded by low light or adverse weather.

These findings translate into clear practical implications for system engineers, summarized in Table~\ref{Table-3}. The results highlight a fundamental performance-efficiency trade-off. For current in-vehicle applications requiring real-time feedback (\textgreater10 FPS), an efficient object detector like YOLO represents the most pragmatic and deployable choice, offering strong accuracy in common scenarios at minimal computational cost. However, for future safety-critical applications, such as Level 3+ ADAS or post-incident forensic analysis, where robustness under adverse and rare conditions is paramount, large VLMs emerge as the technological successor. Although their substantial computational demands currently limits its real-time applicability, their superior contextual reasoning and resilience in challenging environments point to the future of human-centric vehicle perception. Continued advances in hardware and model optimization are expected to narrow this gap.

\begin{table}[htbp]
  \centering
  \caption{Summary of Three Paradigms and Their Practical Trade-offs.}
  \label{Table-3}
  \begin{tabularx}{\linewidth}{
    >{\raggedright\arraybackslash}m{2.5cm}
    >{\centering\arraybackslash\hsize=1.00\hsize}X
    >{\centering\arraybackslash\hsize=0.90\hsize}X
    >{\centering\arraybackslash\hsize=1.30\hsize}X
    >{\centering\arraybackslash\hsize=0.85\hsize}X
    >{\centering\arraybackslash\hsize=0.80\hsize}X
    >{\centering\arraybackslash\hsize=1.15\hsize}X
  }
    \toprule
    \textbf{Paradigm} & 
    \textbf{Overall \newline Performance (Macro F1)} & 
    \textbf{Robustness (Night/Rain) (Macro F1)} & 
    \textbf{Small Object Performance (Traffic Light Recall)} & 
    \textbf{Speed (FPS)} & 
    \textbf{VRAM (GB)} & 
    \textbf{Primary Disadvantage} \\
    \midrule
    Detection-Based & \ \ Good\ \ \, (0.87) & High ($\ge 0.81$) & Weak (0.57) & Real-Time (36.1) & Low (1.53) & Missed detections for small objects \\
    \midrule
    Segmentation-Assisted & Poor ($< 0.35$) & Low ($\le 0.45$) & Poor ($\le 0.20$) & Near Real-Time (10.3) & High (9.20) & ``Part-versus-whole'' semantic gap \\
    \midrule
    VLM-Based & Scale-dependent (0.47--0.85) & Scale-dependent (0.40--0.88) & Scale-dependent (0.54--0.82) & Slow (0.07) & Very High (44.97) & High computational cost \\
    \bottomrule
  \end{tabularx}
\end{table}

\section{CONCLUSIONS}\label{conclusions}

This paper presented a systematic, cross-paradigm evaluation of methods for identifying the semantic object at a driver's point of gaze. The key findings are summarized as follows:

\begin{enumerate}
\def\labelenumi{\arabic{enumi}.}
\item
  Paradigm Effectiveness: Direct object detection (YOLOv13) and large-scale Vision-Language Model (Qwen2.5-VL-32b) paradigms were found to be the most effective approaches for the task. In contrast, the segmentation-assisted paradigm was shown to be fundamentally limited by a ``part-versus-whole'' semantic gap.
\item
  Performance-Efficiency Trade-off: A clear trade-off was established between the two leading paradigms. YOLOv13 offers exceptional real-time efficiency and is the most pragmatic choice for current in-vehicle deployment. Qwen2.5-VL-32b, while computationally demanding, provides superior robustness and contextual understanding, particularly for small objects (e.g., Traffic Light) and under adverse conditions (e.g., nighttime), positioning it as a promising candidate for future safety-critical applications.
\item
  Primary Failure Mode: Across all models, performance degradation in challenging conditions was driven primarily by missed detections (a recall problem). The large VLM (Qwen2.5-VL-32b) demonstrates consistently higher recall under these conditions, highlighting the advantage of its holistic and context-aware perception.
\end{enumerate}

Together, these findings provide a roadmap for the development of next-generation human-aware intelligent vehicle systems. Nonetheless, several limitations point to directions for future work. The benchmark dataset, while diverse, is limited in scale and relies on simulated gaze points that do not reflect the noise characteristics of real eye-tracking hardware. Furthermore, the current analysis is frame-based and does not exploit temporal context. Future research should therefore validate these findings using real-world gaze data, investigating hybrid architectures that balance efficiency and robustness, and extend VLM-based approaches to video inputs to leverage temporal information.



\section*{Data Availability}\label{data-availability}

The dataset is publicly accessible at \url{http://bdd-data.berkeley.edu}


The authors declare no competing interests.

\bibliographystyle{unsrt}  
\bibliography{refs}

@article{ahlst2021EyeTrackingDriver,
  title = {Eye {{Tracking}} in {{Driver Attention Research}}—{{How Gaze Data Interpretations Influence What We Learn}}},
  author = {Ahlström, Christer and Kircher, Katja and Nyström, Marcus and Wolfe, Benjamin},
  date = {2021-12-08},
  journaltitle = {Frontiers in Neuroergonomics},
  shortjournal = {Front. Neuroergonomics},
  volume = {2},
  pages = {778043},
  publisher = {Frontiers},
  issn = {2673-6195},
  doi = {10.3389/fnrgo.2021.778043},
  url = {https://www.frontiersin.org/journals/neuroergonomics/articles/10.3389/fnrgo.2021.778043/full},
  urldate = {2025-10-02},
  langid = {english}
}

@online{bai2023QwenVLVersatileVisionLanguage,
  title = {Qwen-{{VL}}: {{A Versatile Vision-Language Model}} for {{Understanding}}, {{Localization}}, {{Text Reading}}, and {{Beyond}}},
  shorttitle = {Qwen-{{VL}}},
  author = {Bai, Jinze and Bai, Shuai and Yang, Shusheng and Wang, Shijie and Tan, Sinan and Wang, Peng and Lin, Junyang and Zhou, Chang and Zhou, Jingren},
  date = {2023-10-13},
  eprint = {2308.12966},
  eprinttype = {arXiv},
  eprintclass = {cs},
  doi = {10.48550/arXiv.2308.12966},
  url = {http://arxiv.org/abs/2308.12966},
  urldate = {2025-10-02},
  pubstate = {prepublished}
}

@article{dong2011DriverInattentionMonitoring,
  title = {Driver {{Inattention Monitoring System}} for {{Intelligent Vehicles}}: {{A Review}}},
  shorttitle = {Driver {{Inattention Monitoring System}} for {{Intelligent Vehicles}}},
  author = {Dong, Yanchao and Hu, Zhencheng and Uchimura, Keiichi and Murayama, Nobuki},
  date = {2011-06},
  journaltitle = {IEEE Transactions on Intelligent Transportation Systems},
  volume = {12},
  number = {2},
  pages = {596--614},
  issn = {1558-0016},
  doi = {10.1109/TITS.2010.2092770},
  url = {https://ieeexplore.ieee.org/abstract/document/5665773},
  urldate = {2025-10-17}
}

@article{elhen2025VisionLanguageModelsAutonomous,
  title = {Vision-{{Language Models}} for {{Autonomous Driving}}: {{CLIP-Based Dynamic Scene Understanding}}},
  shorttitle = {Vision-{{Language Models}} for {{Autonomous Driving}}},
  author = {Elhenawy, Mohammed and Ashqar, Huthaifa I. and Rakotonirainy, Andry and Alhadidi, Taqwa I. and Jaber, Ahmed and Tami, Mohammad Abu},
  date = {2025-01},
  journaltitle = {Electronics},
  volume = {14},
  number = {7},
  pages = {1282},
  publisher = {Multidisciplinary Digital Publishing Institute},
  issn = {2079-9292},
  doi = {10.3390/electronics14071282},
  url = {https://www.mdpi.com/2079-9292/14/7/1282},
  urldate = {2025-10-02},
  langid = {english}
}

@article{feng2024Humancentreddesignnext,
  title = {Human-Centred Design of next Generation Transportation Infrastructure with Connected and Automated Vehicles: A System-of-Systems Perspective},
  shorttitle = {Human-Centred Design of next Generation Transportation Infrastructure with Connected and Automated Vehicles},
  author = {Feng, Yiheng and Chen, Yunfeng and Zhang, Jiansong and Tian, Chi and Ren, Ran and Han, Tianfang and Proctor, Robert W.},
  date = {2024-05-03},
  journaltitle = {Theoretical Issues in Ergonomics Science},
  volume = {25},
  number = {3},
  pages = {287--315},
  publisher = {Taylor \& Francis},
  issn = {1463-922X},
  doi = {10.1080/1463922X.2023.2182003},
  url = {https://doi.org/10.1080/1463922X.2023.2182003},
  urldate = {2025-10-17}
}

@article{garci2021AssessmentInfluenceTechnologyBased,
  title = {Assessment of the {{Influence}} of {{Technology-Based Distracted Driving}} on {{Drivers}}’ {{Infractions}} and {{Their Subsequent Impact}} on {{Traffic Accidents Severity}}},
  author = {García-Herrero, Susana and Febres, Juan Diego and Boulagouas, Wafa and Gutiérrez, José Manuel and Mariscal Saldaña, Miguel Ángel},
  date = {2021-07-04},
  journaltitle = {International Journal of Environmental Research and Public Health},
  shortjournal = {Int J Environ Res Public Health},
  volume = {18},
  number = {13},
  eprint = {34281092},
  eprinttype = {pubmed},
  pages = {7155},
  issn = {1661-7827},
  doi = {10.3390/ijerph18137155},
  url = {https://pmc.ncbi.nlm.nih.gov/articles/PMC8297255/},
  urldate = {2025-10-02},
  pmcid = {PMC8297255}
}

@article{guo2017effectsagecrash,
  title = {The Effects of Age on Crash Risk Associated with Driver Distraction},
  author = {Guo, Feng and Klauer, Sheila G and Fang, Youjia and Hankey, Jonathan M and Antin, Jonathan F and Perez, Miguel A and Lee, Suzanne E and Dingus, Thomas A},
  date = {2017-02-01},
  journaltitle = {International Journal of Epidemiology},
  shortjournal = {Int J Epidemiol},
  volume = {46},
  number = {1},
  pages = {258--265},
  issn = {0300-5771},
  doi = {10.1093/ije/dyw234},
  url = {https://doi.org/10.1093/ije/dyw234},
  urldate = {2025-10-02}
}

@article{haghz2024ClassifyingOlderDrivers,
  title = {Classifying {{Older Drivers}}’ {{Gaze Behaviour}} during {{Automated}} versus {{Non-Automated Driving}}: {{A Preliminary Step}} towards {{Detecting Mode Confusion}}},
  shorttitle = {Classifying {{Older Drivers}}’ {{Gaze Behaviour}} during {{Automated}} versus {{Non-Automated Driving}}},
  author = {Haghzare, Shabnam and Campos, Jennifer L. and Mihailidis, Alex},
  date = {2024-01-17},
  journaltitle = {International Journal of Human–Computer Interaction},
  volume = {40},
  number = {2},
  pages = {241--254},
  publisher = {Taylor \& Francis},
  issn = {1044-7318},
  doi = {10.1080/10447318.2022.2112933},
  url = {https://doi.org/10.1080/10447318.2022.2112933},
  urldate = {2025-10-02}
}

@article{hazza2024SegmentAnythingReview,
  title = {Segment {{Anything}}: {{A Review}}},
  shorttitle = {Segment {{Anything}}},
  author = {Hazzaa, Firas and Udoidiong, Innocent and Qashou, Akram and Yousef, Sufian},
  date = {2024-10-18},
  journaltitle = {Mesopotamian Journal of Computer Science},
  volume = {2024},
  pages = {150--161},
  issn = {2958-6631},
  doi = {10.58496/MJCSC/2024/012},
  url = {https://mesopotamian.press/journals/index.php/cs/article/view/565},
  urldate = {2025-10-02},
  langid = {english}
}

@article{hersl2003Lookedbutfailedtoseeerrorstraffic,
  title = {Looked-but-Failed-to-See-Errors in Traffic},
  author = {Herslund, Mai-Britt and Jørgensen, Niels O},
  date = {2003-11-01},
  journaltitle = {Accident Analysis \& Prevention},
  shortjournal = {Accident Analysis \& Prevention},
  volume = {35},
  number = {6},
  pages = {885--891},
  issn = {0001-4575},
  doi = {10.1016/S0001-4575(02)00095-7},
  url = {https://www.sciencedirect.com/science/article/pii/S0001457502000957},
  urldate = {2025-10-02}
}

@article{hoffm2021RealTimeAdaptiveObject,
  title = {Real-{{Time Adaptive Object Detection}} and {{Tracking}} for {{Autonomous Vehicles}}},
  author = {Hoffmann, João Eduardo and Tosso, Hilkija Gaïus and Santos, Max Mauro Dias and Justo, João Francisco and Malik, Asad Waqar and Rahman, Anis Ur},
  date = {2021-09},
  journaltitle = {IEEE Transactions on Intelligent Vehicles},
  volume = {6},
  number = {3},
  pages = {450--459},
  issn = {2379-8904},
  doi = {10.1109/TIV.2020.3037928},
  url = {https://ieeexplore.ieee.org/abstract/document/9259200},
  urldate = {2025-10-17}
}

@article{kenne2013InattentionalBlindnessSimulated,
  title = {Inattentional {{Blindness}} in a {{Simulated Driving Task}}},
  author = {Kennedy, Kellie D. and Bliss, James P.},
  date = {2013-09-01},
  journaltitle = {Proceedings of the Human Factors and Ergonomics Society Annual Meeting},
  volume = {57},
  number = {1},
  pages = {1899--1903},
  publisher = {SAGE Publications Inc},
  issn = {1071-1813},
  doi = {10.1177/1541931213571423},
  url = {https://doi.org/10.1177/1541931213571423},
  urldate = {2025-10-02},
  langid = {english}
}

@inproceedings{keska2025EvaluatingMultimodalVisionLanguage,
  title = {Evaluating {{Multimodal Vision-Language Model Prompting Strategies}} for {{Visual Question Answering}} in {{Road Scene Understanding}}},
  booktitle = {2025 {{IEEE}}/{{CVF Winter Conference}} on {{Applications}} of {{Computer Vision Workshops}} ({{WACVW}})},
  author = {Keskar, Aryan and Perisetla, Srinivasa and Greer, Ross},
  date = {2025-02},
  pages = {937--946},
  issn = {2690-621X},
  doi = {10.1109/WACVW65960.2025.00115},
  url = {https://ieeexplore.ieee.org/document/10972502},
  urldate = {2025-10-02},
  eventtitle = {2025 {{IEEE}}/{{CVF Winter Conference}} on {{Applications}} of {{Computer Vision Workshops}} ({{WACVW}})}
}

@article{khan2019GazeEyeTracking,
  title = {Gaze and {{Eye Tracking}}: {{Techniques}} and {{Applications}} in {{ADAS}}},
  shorttitle = {Gaze and {{Eye Tracking}}},
  author = {Khan, Muhammad Qasim and Lee, Sukhan},
  date = {2019-12-14},
  journaltitle = {Sensors (Basel, Switzerland)},
  shortjournal = {Sensors (Basel)},
  volume = {19},
  number = {24},
  eprint = {31847432},
  eprinttype = {pubmed},
  pages = {5540},
  issn = {1424-8220},
  doi = {10.3390/s19245540},
  url = {https://pmc.ncbi.nlm.nih.gov/articles/PMC6960643/},
  urldate = {2025-10-02},
  pmcid = {PMC6960643}
}

@article{kim2025SustainableRealTimeDriver,
  title = {Sustainable {{Real-Time Driver Gaze Monitoring}} for {{Enhancing Autonomous Vehicle Safety}}},
  author = {Kim, Jong-Bae},
  date = {2025-01},
  journaltitle = {Sustainability},
  volume = {17},
  number = {9},
  pages = {4114},
  publisher = {Multidisciplinary Digital Publishing Institute},
  issn = {2071-1050},
  doi = {10.3390/su17094114},
  url = {https://www.mdpi.com/2071-1050/17/9/4114},
  urldate = {2025-10-02},
  langid = {english}
}

@article{klaue2015EffectSecondaryTask,
  title = {The {{Effect}} of {{Secondary Task Engagement}} on {{Adolescents}}' {{Driving Performance}} and {{Crash Risk}}},
  author = {Klauer, Sheila G. and Ehsani, Johnathon P. and McGehee, Daniel V. and Manser, Michael},
  date = {2015-07-01},
  journaltitle = {Journal of Adolescent Health},
  shortjournal = {Journal of Adolescent Health},
  series = {Exploring {{Teen Driver Safety}} and {{Crash Risk}}: {{State}} of the {{Research}}},
  volume = {57},
  pages = {S36-S43},
  issn = {1054-139X},
  doi = {10.1016/j.jadohealth.2015.03.014},
  url = {https://www.sciencedirect.com/science/article/pii/S1054139X15001214},
  urldate = {2025-10-18},
  issue = {1, Supplement}
}

@article{kumar2023ObjectDetectionAdverse,
  title = {Object {{Detection}} in {{Adverse Weather}} for {{Autonomous Driving}} through {{Data Merging}} and {{YOLOv8}}},
  author = {Kumar, Debasis and Muhammad, Naveed},
  date = {2023-01},
  journaltitle = {Sensors},
  volume = {23},
  number = {20},
  pages = {8471},
  publisher = {Multidisciplinary Digital Publishing Institute},
  issn = {1424-8220},
  doi = {10.3390/s23208471},
  url = {https://www.mdpi.com/1424-8220/23/20/8471},
  urldate = {2025-10-02},
  langid = {english}
}

@article{ledez2021ImplementingGazeTracking,
  title = {Implementing a {{Gaze Tracking Algorithm}} for {{Improving Advanced Driver Assistance Systems}}},
  author = {Ledezma, Agapito and Zamora, Víctor and Sipele, Óscar and Sesmero, M. Paz and Sanchis, Araceli},
  date = {2021-01},
  journaltitle = {Electronics},
  volume = {10},
  number = {12},
  pages = {1480},
  publisher = {Multidisciplinary Digital Publishing Institute},
  issn = {2079-9292},
  doi = {10.3390/electronics10121480},
  url = {https://www.mdpi.com/2079-9292/10/12/1480},
  urldate = {2025-10-02},
  langid = {english}
}

@online{lei2025YOLOv13RealTimeObject,
  title = {{{YOLOv13}}: {{Real-Time Object Detection}} with {{Hypergraph-Enhanced Adaptive Visual Perception}}},
  shorttitle = {{{YOLOv13}}},
  author = {Lei, Mengqi and Li, Siqi and Wu, Yihong and Hu, Han and Zhou, You and Zheng, Xinhu and Ding, Guiguang and Du, Shaoyi and Wu, Zongze and Gao, Yue},
  date = {2025-09-05},
  eprint = {2506.17733},
  eprinttype = {arXiv},
  eprintclass = {cs},
  doi = {10.48550/arXiv.2506.17733},
  url = {http://arxiv.org/abs/2506.17733},
  urldate = {2025-10-18},
  pubstate = {prepublished}
}

@online{lin2015MicrosoftCOCOCommon,
  title = {Microsoft {{COCO}}: {{Common Objects}} in {{Context}}},
  shorttitle = {Microsoft {{COCO}}},
  author = {Lin, Tsung-Yi and Maire, Michael and Belongie, Serge and Bourdev, Lubomir and Girshick, Ross and Hays, James and Perona, Pietro and Ramanan, Deva and Zitnick, C. Lawrence and Dollár, Piotr},
  date = {2015-02-21},
  eprint = {1405.0312},
  eprinttype = {arXiv},
  eprintclass = {cs},
  doi = {10.48550/arXiv.1405.0312},
  url = {http://arxiv.org/abs/1405.0312},
  urldate = {2025-10-18},
  pubstate = {prepublished}
}

@article{liu2025YOLOFDCLImprovedYOLOv8,
  title = {{{YOLO-FDCL}}: {{Improved YOLOv8}} for {{Driver Fatigue Detection}} in {{Complex Lighting Conditions}}},
  shorttitle = {{{YOLO-FDCL}}},
  author = {Liu, Genchao and Wu, Kun and Lan, Wei and Wu, Yunjie},
  date = {2025-08-06},
  journaltitle = {Sensors (Basel, Switzerland)},
  shortjournal = {Sensors (Basel)},
  volume = {25},
  number = {15},
  eprint = {40807995},
  eprinttype = {pubmed},
  pages = {4832},
  issn = {1424-8220},
  doi = {10.3390/s25154832},
  url = {https://pmc.ncbi.nlm.nih.gov/articles/PMC12349288/},
  urldate = {2025-10-02},
  pmcid = {PMC12349288}
}

@inproceedings{maity2021FasterRCNNYOLO,
  title = {Faster {{R-CNN}} and {{YOLO}} Based {{Vehicle}} Detection: {{A Survey}}},
  shorttitle = {Faster {{R-CNN}} and {{YOLO}} Based {{Vehicle}} Detection},
  booktitle = {2021 5th {{International Conference}} on {{Computing Methodologies}} and {{Communication}} ({{ICCMC}})},
  author = {Maity, Madhusri and Banerjee, Sriparna and Sinha Chaudhuri, Sheli},
  date = {2021-04},
  pages = {1442--1447},
  doi = {10.1109/ICCMC51019.2021.9418274},
  url = {https://ieeexplore.ieee.org/abstract/document/9418274},
  urldate = {2025-10-18},
  eventtitle = {2021 5th {{International Conference}} on {{Computing Methodologies}} and {{Communication}} ({{ICCMC}})}
}

@inproceedings{marda2019ResolvingTargetAmbiguity,
  title = {Resolving {{Target Ambiguity}} in {{3D Gaze Interaction}} through {{VOR Depth Estimation}}},
  booktitle = {Proceedings of the 2019 {{CHI Conference}} on {{Human Factors}} in {{Computing Systems}}},
  author = {Mardanbegi, Diako and Langlotz, Tobias and Gellersen, Hans},
  date = {2019-05-02},
  series = {{{CHI}} '19},
  pages = {1--12},
  publisher = {Association for Computing Machinery},
  location = {New York, NY, USA},
  doi = {10.1145/3290605.3300842},
  url = {https://dl.acm.org/doi/10.1145/3290605.3300842},
  urldate = {2025-10-02},
  isbn = {978-1-4503-5970-2}
}

@article{mirza2023SmallObjectDetection,
  title = {Small {{Object Detection}} and {{Tracking}}: {{A Comprehensive Review}}},
  shorttitle = {Small {{Object Detection}} and {{Tracking}}},
  author = {Mirzaei, Behzad and Nezamabadi-pour, Hossein and Raoof, Amir and Derakhshani, Reza},
  date = {2023-01},
  journaltitle = {Sensors},
  volume = {23},
  number = {15},
  pages = {6887},
  publisher = {Multidisciplinary Digital Publishing Institute},
  issn = {1424-8220},
  doi = {10.3390/s23156887},
  url = {https://www.mdpi.com/1424-8220/23/15/6887},
  urldate = {2025-10-02},
  langid = {english}
}

@report{natio2025Distracteddriving2023,
  title = {Distracted Driving in 2023},
  author = {{National Center for Statistics and Analysis}},
  date = {2025-04},
  number = {Report No. DOT HS 813 703},
  institution = {National Highway Traffic Safety Administration},
  location = {Washington, DC},
  langid = {english}
}

@report{natio2025EarlyEstimateMotor,
  title = {Early {{Estimate}} of {{Motor Vehicle Traffic Fatalities}} in 2024},
  author = {{National Center for Statistics and Analysis}},
  date = {2025-04},
  number = {DOT HS 813 710},
  institution = {National Highway Traffic Safety Administration},
  location = {Washington, DC},
  langid = {english}
}

@online{ravi2024SAM2Segment,
  title = {{{SAM}} 2: {{Segment Anything}} in {{Images}} and {{Videos}}},
  shorttitle = {{{SAM}} 2},
  author = {Ravi, Nikhila and Gabeur, Valentin and Hu, Yuan-Ting and Hu, Ronghang and Ryali, Chaitanya and Ma, Tengyu and Khedr, Haitham and Rädle, Roman and Rolland, Chloe and Gustafson, Laura and Mintun, Eric and Pan, Junting and Alwala, Kalyan Vasudev and Carion, Nicolas and Wu, Chao-Yuan and Girshick, Ross and Dollár, Piotr and Feichtenhofer, Christoph},
  date = {2024-10-28},
  eprint = {2408.00714},
  eprinttype = {arXiv},
  eprintclass = {cs},
  doi = {10.48550/arXiv.2408.00714},
  url = {http://arxiv.org/abs/2408.00714},
  urldate = {2025-10-18},
  pubstate = {prepublished}
}

@article{simon2014KeepYourEyes,
  title = {Keep {{Your Eyes}} on the {{Road}}: {{Young Driver Crash Risk Increases According}} to {{Duration}} of {{Distraction}}},
  shorttitle = {Keep {{Your Eyes}} on the {{Road}}},
  author = {Simons-Morton, Bruce G. and Guo, Feng and Klauer, Sheila G. and Ehsani, Johnathon P. and Pradhan, Anuj K.},
  date = {2014-05-01},
  journaltitle = {Journal of Adolescent Health},
  shortjournal = {Journal of Adolescent Health},
  series = {Driver {{Distraction}}: {{A Perennial}} but {{Preventable Public Health Threat}} to {{Adolescents}}},
  volume = {54},
  pages = {S61-S67},
  issn = {1054-139X},
  doi = {10.1016/j.jadohealth.2013.11.021},
  url = {https://www.sciencedirect.com/science/article/pii/S1054139X13007799},
  urldate = {2025-10-18},
  issue = {5, Supplement}
}

@article{tammi2025Quantifyingtrackingquality,
  title = {Quantifying Tracking Quality during Occlusion with an Integrated Gaze Metric Anchored to Task Performance},
  author = {Tammi, Tuisku and Pekkanen, Jami and Cowley, Benjamin Ultan and Lappi, Otto},
  date = {2025-08-29},
  journaltitle = {Scientific Reports},
  shortjournal = {Sci Rep},
  volume = {15},
  eprint = {40883515},
  eprinttype = {pubmed},
  pages = {31858},
  issn = {2045-2322},
  doi = {10.1038/s41598-025-17519-8},
  url = {https://pmc.ncbi.nlm.nih.gov/articles/PMC12397211/},
  urldate = {2025-10-02},
  pmcid = {PMC12397211}
}

@inproceedings{tan2021EfficientNetV2SmallerModels,
  title = {{{EfficientNetV2}}: {{Smaller Models}} and {{Faster Training}}},
  shorttitle = {{{EfficientNetV2}}},
  booktitle = {Proceedings of the 38th {{International Conference}} on {{Machine Learning}}},
  author = {Tan, Mingxing and Le, Quoc},
  date = {2021-07-01},
  pages = {10096--10106},
  publisher = {PMLR},
  issn = {2640-3498},
  url = {https://proceedings.mlr.press/v139/tan21a.html},
  urldate = {2025-10-18},
  eventtitle = {International {{Conference}} on {{Machine Learning}}},
  langid = {english}
}

@inproceedings{tonin2023ObjectawareGazeTarget,
  title = {Object-Aware {{Gaze Target Detection}}},
  booktitle = {2023 {{IEEE}}/{{CVF International Conference}} on {{Computer Vision}} ({{ICCV}})},
  author = {Tonini, Francesco and Dall’Asen, Nicola and Beyan, Cigdem and Ricci, Elisa},
  date = {2023-10},
  pages = {21803--21812},
  issn = {2380-7504},
  doi = {10.1109/ICCV51070.2023.01998},
  url = {https://ieeexplore.ieee.org/document/10377304},
  urldate = {2025-10-18},
  eventtitle = {2023 {{IEEE}}/{{CVF International Conference}} on {{Computer Vision}} ({{ICCV}})}
}

@article{walke2019Gazebehaviourelectrodermal,
  title = {Gaze Behaviour and Electrodermal Activity: {{Objective}} Measures of Drivers’ Trust in Automated Vehicles},
  shorttitle = {Gaze Behaviour and Electrodermal Activity},
  author = {Walker, F. and Wang, J. and Martens, M. H. and Verwey, W. B.},
  date = {2019-07-01},
  journaltitle = {Transportation Research Part F: Traffic Psychology and Behaviour},
  shortjournal = {Transportation Research Part F: Traffic Psychology and Behaviour},
  volume = {64},
  pages = {401--412},
  issn = {1369-8478},
  doi = {10.1016/j.trf.2019.05.021},
  url = {https://www.sciencedirect.com/science/article/pii/S1369847818306703},
  urldate = {2025-10-17}
}

@article{wang2023DriverAttentionDetection,
  title = {Driver {{Attention Detection Based}} on {{Improved YOLOv5}}},
  author = {Wang, Zhongzhou and Yao, Keming and Guo, Fuao},
  date = {2023-01},
  journaltitle = {Applied Sciences},
  volume = {13},
  number = {11},
  pages = {6645},
  publisher = {Multidisciplinary Digital Publishing Institute},
  issn = {2076-3417},
  doi = {10.3390/app13116645},
  url = {https://www.mdpi.com/2076-3417/13/11/6645},
  urldate = {2025-10-02},
  langid = {english}
}

@inproceedings{wang2025LearningVisualGrounding,
  title = {Learning {{Visual Grounding}} from {{Generative Vision}} and {{Language Model}}},
  booktitle = {2025 {{IEEE}}/{{CVF Winter Conference}} on {{Applications}} of {{Computer Vision}} ({{WACV}})},
  author = {Wang, Shijie and Kim, Dahun and Taalimi, Ali and Sun, Chen and Kuo, Weicheng},
  date = {2025-02},
  pages = {8057--8067},
  issn = {2642-9381},
  doi = {10.1109/WACV61041.2025.00782},
  url = {https://ieeexplore.ieee.org/abstract/document/10943391},
  urldate = {2025-10-02},
  eventtitle = {2025 {{IEEE}}/{{CVF Winter Conference}} on {{Applications}} of {{Computer Vision}} ({{WACV}})}
}

@article{winla2019Usingtelematicsdata,
  title = {Using Telematics Data to Find Risky Driver Behaviour},
  author = {Winlaw, Manda and Steiner, Stefan H. and MacKay, R. Jock and Hilal, Allaa R.},
  date = {2019-10-01},
  journaltitle = {Accident Analysis \& Prevention},
  shortjournal = {Accident Analysis \& Prevention},
  volume = {131},
  pages = {131--136},
  issn = {0001-4575},
  doi = {10.1016/j.aap.2019.06.003},
  url = {https://www.sciencedirect.com/science/article/pii/S0001457519304956},
  urldate = {2025-10-02}
}

@article{wolfe2022Normalblindnesswhen,
  title = {Normal Blindness: When We {{Look}} but {{Fail To See}}},
  shorttitle = {Normal Blindness},
  author = {Wolfe, Jeremy M and Kosovicheva, Anna and Wolfe, Benjamin},
  date = {2022-09},
  journaltitle = {Trends in cognitive sciences},
  shortjournal = {Trends Cogn Sci},
  volume = {26},
  number = {9},
  eprint = {35872002},
  eprinttype = {pubmed},
  pages = {809--819},
  issn = {1364-6613},
  doi = {10.1016/j.tics.2022.06.006},
  url = {https://pmc.ncbi.nlm.nih.gov/articles/PMC9378609/},
  urldate = {2025-10-02},
  pmcid = {PMC9378609}
}

@report{world2023GlobalStatusReport,
  title = {Global {{Status Report}} on {{Road Safety}} 2023},
  author = {{World Health Organization}},
  date = {2023},
  institution = {World Health Organization},
  location = {Geneva},
  langid = {english}
}

@article{youss2022Trafficsignclassification,
  title = {Traffic Sign Classification Using {{CNN}} and Detection Using Faster-{{RCNN}} and {{YOLOV4}}},
  author = {Youssouf, Njayou},
  date = {2022-12-01},
  journaltitle = {Heliyon},
  shortjournal = {Heliyon},
  volume = {8},
  number = {12},
  pages = {e11792},
  issn = {2405-8440},
  doi = {10.1016/j.heliyon.2022.e11792},
  url = {https://www.sciencedirect.com/science/article/pii/S2405844022030808},
  urldate = {2025-10-02}
}

@online{yu2020BDD100KDiverseDriving,
  title = {{{BDD100K}}: {{A Diverse Driving Dataset}} for {{Heterogeneous Multitask Learning}}},
  shorttitle = {{{BDD100K}}},
  author = {Yu, Fisher and Chen, Haofeng and Wang, Xin and Xian, Wenqi and Chen, Yingying and Liu, Fangchen and Madhavan, Vashisht and Darrell, Trevor},
  date = {2020-04-08},
  eprint = {1805.04687},
  eprinttype = {arXiv},
  eprintclass = {cs},
  doi = {10.48550/arXiv.1805.04687},
  url = {http://arxiv.org/abs/1805.04687},
  urldate = {2025-10-18},
  pubstate = {prepublished}
}

@article{yuan2024Principlesapplicationsadvancements,
  title = {Principles, Applications, and Advancements of the {{Segment Anything Model}}},
  author = {Yuan, Zhongkai},
  date = {2024-03-28},
  journaltitle = {Applied and Computational Engineering},
  volume = {53},
  pages = {73--78},
  issn = {2755-273X, 2755-2721},
  doi = {10.54254/2755-2721/53/20241270},
  url = {https://www.ewadirect.com/proceedings/ace/article/view/10924},
  urldate = {2025-10-02},
  langid = {english}
}

@article{zhao2024Roadsurfacesemantic,
  title = {Road Surface Semantic Segmentation for Autonomous Driving},
  author = {Zhao, Huaqi and Wang, Su and Peng, Xiang and Pan, Jeng-Shyang and Wang, Rui and Liu, Xiaomin},
  date = {2024-09-25},
  journaltitle = {PeerJ Computer Science},
  shortjournal = {PeerJ Comput. Sci.},
  volume = {10},
  pages = {e2250},
  publisher = {PeerJ Inc.},
  issn = {2376-5992},
  doi = {10.7717/peerj-cs.2250},
  url = {https://peerj.com/articles/cs-2250},
  urldate = {2025-10-02},
  langid = {english}
}

@article{zhu2024DrivingFutureExploring,
  title = {Driving {{Towards}} the {{Future}}: {{Exploring Human-Centered Design}} and {{Experiment}} of {{Glazing Projection Display Systems}} for {{Autonomous Vehicles}}},
  shorttitle = {Driving {{Towards}} the {{Future}}},
  author = {Zhu, Yancong and Geng, Yunke and Huang, Ruonan and Zhang, Xiaonan and Wang, Lu and Liu, Wei},
  date = {2024-08-02},
  journaltitle = {International Journal of Human–Computer Interaction},
  volume = {40},
  number = {15},
  pages = {4087--4102},
  publisher = {Taylor \& Francis},
  issn = {1044-7318},
  doi = {10.1080/10447318.2023.2209836},
  url = {https://doi.org/10.1080/10447318.2023.2209836},
  urldate = {2025-10-02}
}

\end{document}